# The inattentional gap in task conditioned AI models that omit otherwise reportable safety critical signals

**Kwan Soo Shin**

PolymathMinds Lab, Asan, Republic of Korea

Corresponding author: Kwan Soo Shin, sshin@pmminds.ai (ORCID: 0009-0001-5799-7556)

---

## Abstract

AI in radiology and other safety-critical workflows is evaluated on the hazards it is told to find, yet harm arises disproportionately from hazards no one specified. We show that conditioning a language or vision model on a narrow task suppresses its reporting of co-present, safety-critical signals it can otherwise report, a behavioral analogue of human inattentional blindness. Across radiology text scenarios and thoracic-image vision tasks, ordinary focused instructions suppressed reporting by up to 0.92; the gap ranged from minimal to complete across seven models, did not vary monotonically with scale, and persisted in a reasoning model, while one flagship model showed a robust safety-reporting override. We term this dissociation the Inattentional Gap: a system can score near-perfectly on specified hazards while omitting co-present safety-critical hazards. In a 24-scenario probe, an independent open-ended critic restored every omitted finding. We propose reporting-complete evaluation as an admission criterion for safety-critical deployment.

## Introduction

Accidents happen where they are not expected. The hazards an operator anticipates are, by construction, the hazards a system is built and tested to detect; the residual risk lives in the events outside that specification. This asymmetry is the organizing intuition of this paper, and it has a precise correlate in human cognition. In the “invisible gorilla” paradigm, observers counting basketball passes fail to notice a person in a gorilla suit walking through the scene (1), and the effect persists in domain experts: 83% of radiologists searching a chest CT for lung nodules failed to report a matchbook-sized gorilla composited into the image, many having fixated directly on it (2). Inattentional blindness is not simply a failure of acuity. It is a failure of report under an attentional set: the observer looks, but the task has narrowed what counts as relevant (3).

Large language and vision-language models are now routinely conditioned on narrow tasks and deployed as perceptual or interpretive front-ends in safety-critical settings. A radiology assistant is asked to assess a nodule; a driving system is asked to track the lead vehicle; a screening model is asked to flag a specified threat. The question we pose is whether task-conditioning produces, in these systems, a functional analogue of inattentional blindness: suppression of safety-critical information that the model can otherwise report under an unconstrained instruction.

This question sits at the intersection of two literatures. The first, machine psychology, treats models as participants in cognitive experiments and has documented that language models reproduce, and sometimes shed, human reasoning biases (4, 5). The second concerns the limits of machine perception and visual-language reporting, including evidence that vision-language models miss visually salient elements (6) and that behavioral parity with humans need not imply mechanistic equivalence (7, 8). What has not been established is whether an attentional-set manipulation, the defining cause of inattentional blindness in humans, suppresses safety-critical reporting in AI systems, and whether this effect holds across modalities, domains, and model scales.

We make four contributions. First, we formalize the Inattentional Gap: a divergence between what a model reports under a narrow task and what the same model reports when unconstrained. Second, we demonstrate the gap across two modalities, in language and vision, within a safety-critical clinical domain and with generalization to a second domain (Supplementary Note 1), using a controlled attentional-set manipulation and an adjudication procedure that separates task-induced omission from mere capability failure. Third, we show that the gap is not reliably eliminated by model scale, persists in a reasoning model, and varies with modality, task load, signal salience, and model family; we further implicate output scope as the best-supported proximal behavioral account and probe its behavioral status, finding System-1-style task capture with no reliable architecture-general oversight behavior in the systems sampled here. Fourth, we demonstrate a validated mitigation: an external open-ended critic, run as an explicit second process, restored every omitted finding in our probe, turning the gap from an observed hazard into an engineering and evaluation target. Together, these results map the conditions under which a task-conditioned AI system omits safety-critical signals it can otherwise report, and what closes the gap.

Let a deployed model be conditioned by a task instruction that designates a target set D_specified, while the same scene also contains co-present, unrequested, safety-critical information D_unspecified. Conventional safety benchmarks measure performance on D_specified; deployment harm often turns on whether D_unspecified is surfaced. The Inattentional Gap denotes the regime in which a model can perform well on the specified target while suppressing report of the unspecified safety-critical signal. Formally, for model $m$ and item $i$, let $R_{m,i}^{\text{open}}$ and $R_{m,i}^{\text{task}}$ indicate whether the safety-critical signal is reported under an unconstrained instruction and a task-conditioned instruction, respectively. The item-level gap is $IG_{m,i} = R_{m,i}^{\text{open}} - R_{m,i}^{\text{task}}$, and the model-level gap is estimated over items for which $R_{m,i}^{\text{open}} = 1$, so that the effect is conditioned on open-condition reportability rather than on a presumed perceptual state (Figure 1).

The construct is behavioral and falsifiable. Its signature is a within-item dissociation: the same model, on the same input, reports a critical signal when unconstrained but omits it when the task instruction narrows the reporting frame. This control distinguishes task-induced omission from a capability deficit. An item whose open condition does not surface the signal cannot support a claim of suppression and is therefore excluded from the gap estimate.

A natural objection is that a model omitting a finding under a narrowing instruction is merely following instructions, not exhibiting an analogue of inattentional blindness. But in the canonical human paradigm, the omission is itself induced by a task instruction: observers miss the gorilla because they were told to count the passes of the white-shirted team (1), and what is noticed is shaped by the attentional set the task imposes, "what you see is what you set" (9). The clinical form is similar: eleven of twelve ophthalmologists missed signs of iron toxicity when the posed question asked them to rule out melanoma (10), and neuroradiologists on a routine read overlooked lesions that the same images yielded to readers instructed to seek them (11) (for a review of inattentional blindness in medicine, see Hults et al. (12)). Instruction-following is therefore not a rival explanation for the Inattentional Gap; it is the channel through which an attentional set is imposed, in humans by instruction or expectation and in models by the prompt. The open-condition recovery provides the machine analogue of the full-attention control: the signal is first shown to be reportable, then shown to be omitted once the task set is imposed.

The construct connects to a broader hypothesis about closed-system reasoning failure: a model solves an open-world problem by reducing it to the closed world its prompt defines, and fails outside that closure. The Inattentional Gap is the report-level instance of that failure.

This closure is not merely a prompt-engineering defect but a measurable risk of task-scoped deployment. To guarantee that no co-present critical signal is ever suppressed, a system would have to analyze every input exhaustively and without scoping, the open condition run universally, a single model reporting everything that might matter on every case. No deployed system can do this without cost. Real deployment is therefore necessarily task-scoped: a chest CT is routed to a nodule detector, a product is cleared for one indication, or a clinician asks only whether a nodule is present. The same scoping that makes AI tractable, certifiable, and affordable is also the condition under which unrequested safety-critical signals can be omitted.

The Inattentional Gap sits at the convergence of two research lineages that, to our knowledge, have not been joined in a within-item, reportability-controlled, safety-critical design in prior machine studies. Its novelty is clearest when read against both. We surveyed 116 prior works across eleven axes (full annotated bibliography, Supplementary Information). The main text cites the lineage anchors that bear directly on the argument, while Figure 6 provides a structured positioning map; the complete annotated bibliography, axis coding, and search rationale are provided in the Supplementary Information.

A human lineage runs from laboratory to clinic. Selective-looking studies (13) led to the formalization of inattentional blindness as the principle that without attention there is no report (3), made vivid by the invisible gorilla (1), and Most et al. (9) identified the operative variable, the attentional set, in the phrase "what you see is what you set." The clinical instantiation is satisfaction of search: detecting one abnormality lowers detection of a co-present second (14), and eye tracking shows that the missed second finding is often fixated, looked at but not reported (15), a pattern now unified across radiology and the laboratory as subsequent-search-miss error (16). The effect survives expertise: 83% of radiologists missed a gorilla composited into a chest scan (2), the miss extends to clinically relevant pathology that experience does not protect against (17), and greater expertise can deepen the blindness by sharpening the attentional set (18). Each of these studies established set-dependent omission in humans; none asked whether the same manipulation suppresses reporting in machines.

A machine lineage treats models as cognitive subjects. From the question of whether machines think like people (19), the machine-psychology programme (20) applied cognitive-experimental tools to language models (4), documenting human-like content effects (21), emergent then alignment-suppressed biases (5), and a dissociation of formal from functional competence (22). A parallel strand argues that what a model "knows" is structured by its task (23), the static precursor to the live, inference-time gating we measure. This lineage probes reasoning and competence; it does not manipulate an attentional set over an otherwise reportable, safety-critical signal. That these models now reach human-level performance on tasks from analogical reasoning to theory of mind (24) sharpens the point: the conditioned omission we report is task-induced omission rather than a capability failure. The two lineages meet at our question.

The claim has direct interlocutors at the intersection of consciousness science, prediction safety, and model behavior. Block's (25) thesis that perceptual consciousness can exceed cognitive access provides the theoretical analogue; here we operationalize only a report-level dissociation in machines, where the prompt narrows the model's usable report bandwidth (26). van Amsterdam et al. (27) showed that accurate prediction models can yield harmful self-fulfilling prophecies; we extend their accuracy-harm decoupling from prediction loops to report under task-conditioning. Park et al. (28) catalogue strategic deception, models that misreport to reach a goal; the Inattentional Gap is its non-strategic counterpart, omission with no apparent goal to hide but a narrowed reporting frame. Mahowald et al. (22) dissociated formal from functional competence, and we add a third dissociation: otherwise reportable signal versus task-conditioned report. Sanchez et al. (29) built a radiologist-AI disagreement pipeline that does not directly target silent omission, the failure mode isolated here.

The nearest machine neighbors each fall one axis short. Vision-language-limit studies, including saliency benchmarks (6), demonstrations that capable models miss obvious features (30), and findings that models fail on representationally confusable inputs (31), ask the model directly and measure capability, whereas we measure suppression of what the model is not asked to report. Object-hallucination benchmarks (32) measure the opposite error, reporting what is absent, while we measure failing to report what is present; the two are complementary axes of safety-relevant reporting error. AbsenceBench (33) tests detection of deliberately removed tokens with a direct query, not set-induced suppression of present, otherwise reportable content under a competing task. Each neighbor stops one axis short.

The deployment-safety literature our thesis ultimately addresses concerns benchmark scores, distribution shift, and specification gaming. Task-conditioning is, in this frame, a prompt-induced shortcut: the model satisfies the stated task and shortcuts past everything else (34), an inference-time instance of two canonical safety problems, negative side effects and evaluation-to-deployment distributional shift (35), and the prompt-level analogue of agents that game a literal specification while violating its intent (36). That aggregate scores hide within-case failures is documented in medical imaging (37), and that evaluation routinely diverges from deployment reality is visible in the regulatory record of cleared medical AI (38). In clinical language systems this divergence has now been named directly: current evaluations have been argued to constitute an evaluation illusion, built on data, tasks, and metrics that fail to capture real-world efficacy (39). The empirical record supports the charge at two levels: a multimodal model can reach expert-level accuracy on curated image challenges while resting on flawed rationales (40), and deployed chatbots answering patient-posed questions omit information that physicians judge critical (41). Each of these studies documents that measured competence overstates deployment safety; none manipulates the attentional set that task-scoped deployment itself imposes, which is the variable isolated here. Our contribution locates this decoupling at the within-item reporting level: specified-target performance on D_specified, reporting coverage on D_unspecified.

The mechanistic reading is constrained by work showing that behavioral parity need not imply mechanistic equivalence (7, 8). Transformers do not instantiate the same human capacity-limited attentional bottleneck (42), so convergence with the human gorilla effect is informative precisely because the mechanism likely differs. We locate the machine-side mechanism in report-conditioning rather than in a human-like perceptual bottleneck. In dual-process terms (43), a narrow prompt induces System-1-style task capture, while the bottom-up salience that a classical model (44) would predict to capture attention can be overridden by the top-down task set.

The consequence is anticipated by the human-factors lineage on the ironies of automation (45): the overseer most likely to trust an accurate system is least able to recover what it silently omits (46). Automation complacency is itself attentional (47), with humans allocating less scrutiny to automated channels, so a model's task-conditioned reporting gate and an operator's complacency can compound; trust decouples from reliability (48), and in medicine automation bias predicts that a clinician may not independently recover an omitted finding (49). The prediction has experimental support: physicians receiving inaccurate advice on chest radiographs performed roughly 40% worse than those receiving accurate advice, whether that advice was labeled as coming from an AI or from a human colleague (50). Two attention gates in series, the model's and the overseer's, define the deployment risk surface.

Mapping the 116 works on two axes, engagement with inattentional-blindness theory and use of a task-conditioning manipulation with a within-item reportability control, the high-engagement quadrant is occupied only by human studies, while machine clusters sit at low-to-mid values on one or both axes (Figure 6). The machine-side high-engagement, high-control cell is not vacant but underdeveloped, and its nearest neighbors sharpen rather than pre-empt the present contribution: a recent benchmark invokes inattentional blindness to show that models under explicit task instructions miss subtle contextual cues in multiple-choice reasoning (51), and multimodal models have been shown to prioritize instruction compliance over surfacing hidden issues under underspecified or misspecified prompts, with explicit prompting revealing the underlying capability and inference-time interventions partially restoring it (52). Neither manipulates the attentional set over a co-present safety-critical signal, and neither establishes item-level reportability before scoring an omission; in the latter, moreover, the hidden issues are defects of the prompt itself (absent objects, contradictions, infeasible requests) rather than signals present in the input. To our knowledge, no prior machine study combines a within-item open-condition reportability control, a task-set manipulation over a co-present safety-critical signal, transfer across language and vision, specified-task competence adjudication, and an independent-critic mitigation test with a specificity control; that combination is the contribution here.

# Results

## Study 1: suppression of reporting in language models

We constructed 64 radiology text scenarios by procedural composition, each pairing a designated target (a pulmonary nodule) with a co-present, unrequested, safety-critical signal (e.g., tension pneumothorax, aortic widening, free subdiaphragmatic air, lytic bone lesions); a parallel set of 36 autonomous-driving scenarios, constructed and analyzed identically, tests generalization beyond the clinical domain and is reported in Supplementary Note 1. Procedural composition reduces the data-contamination concern that limits canonical-vignette studies, since the items are generated from controlled templates rather than copied from known cases. Each scenario was presented to four models (claude-haiku-4-5, gpt-4o-mini, gpt-4o, claude-sonnet-4-6) under three instructions: a high-load focused instruction ("provide a detailed work-up of the nodule"), a strict exclusive instruction ("report only the nodule; do not mention anything else"), and an open instruction ("report all clinically significant findings"). Whether each response reported the critical signal was adjudicated by two independent language-model judges rather than by keyword matching; the author's spot-checks validated judge decisions against response text.

Adjudication showed near-perfect agreement (Cohen $\kappa = 0.895$, $n = 1{,}200$ across the radiology and driving sets); we report the two-judge consensus. The deployment-realistic case is the high-load focused instruction, and there suppression was substantial. Focused-condition report rates were 0.09 (gpt-4o-mini), 0.08 (gpt-4o), 0.48 (sonnet), and 0.53 (haiku), against ~1.00 in the open condition (gaps 0.47 to 0.92; exact McNemar $p =$ 1.4e-17, 3.5e-18, 2.3e-10, and 1.9e-9, respectively): a request as ordinary as a detailed nodule work-up cost the frontier model 92% of the life-threatening findings it otherwise reports. In the driving generalization set the same suppression appeared at weaker magnitude, its more salient hazards resisting suppression (Supplementary Note 1); the contrast identifies a boundary condition: focused-task suppression is larger in higher-load, lower-salience settings (Figure 2).

Pushing the attentional set to its limit, the strict exclusive instruction made suppression complete and universal. Every model reported the critical signal in 0.00 of strict-condition items versus approximately 1.00 of open-condition items (all per-model exact McNemar $p \leq$ 2.2e-19, $N = 64$); in the driving generalization set, strict suppression was strong but partial (Supplementary Note 1). This limit case shows the channel at full strength: the instruction to attend exclusively to the designated target abolished reporting of co-present life-threatening findings that the same models reported when unconstrained.

Two further patterns emerged. The effect did not vary monotonically with scale: gpt-4o, a frontier model, suppressed radiology findings under the focused instruction as strongly as the small gpt-4o-mini (0.08 versus 0.09). The language-model judges also produced stricter estimates than the preliminary lexical parser, indicating that keyword-based measurements understated the focused-condition gap.

## Study 2: suppression of visual report in vision-language models

To test whether the gap extends from language to vision, we turned to the original inattentional-blindness paradigm, in which a demanding primary task competes with report of an unexpected object. We obtained 12 publicly available thoracic images (patient identifiers redacted) and composited a clearly visible but non-anatomical object onto the lung field of each, in the spirit of Drew et al. (2): a subtle translucent disc or a yellow star, varied by side, yielding 48 stimuli. Each image was presented to the four vision-capable models under two instructions: a closed counting task ("carefully count the pairs of ribs"), which consumes attention without invoking diagnosis, and an open instruction ("describe everything you see, including anything unexpected or artificial"). A system message framing the images as synthetic research stimuli eliminated the medical-image refusals that confounded earlier attempts. Object report was again adjudicated by two independent judges, and refusals were excluded.

Task-conditioning suppressed report of the unexpected object in every model (two-judge consensus, Cohen $\kappa = 0.918$, $n = 375$). Under the counting task, report rates were far below the open condition. For the star, closed-condition report rates were 0.00 (gpt-4o), 0.00 (gpt-4o-mini), 0.50 (haiku), and 0.54 (sonnet), against open-condition rates ranging from 0.79 to 0.96. For the disc, closed-condition report rates ranged from 0.00 in both OpenAI models to 0.70 in sonnet, against open-condition rates ranging from 0.67 to 0.96. The strongest effect was observed for gpt-4o on the star condition: the model omitted the star in every counting-task item while reporting it in nearly all open-condition items, a machine analogue of the invisible-gorilla result (exact McNemar $p = 9.5e-7$; Figure 3). Open-condition report was below ceiling for the most subtle objects, reflecting genuine low-salience difficulty; the within-item gap controls for this because it conditions on each model's own open-condition reportability. A post hoc stimulus-integrity audit (Supplementary Note 2) identified two inadvertently duplicated scenes and two non-radiograph images among the twelve; radiograph-only and duplicate-collapsed re-estimates leave the direction of every comparison intact and the closed-condition suppression in both OpenAI models at 0.00 throughout.

Inspection of responses confirmed task-induced omission rather than refusal or a capability failure. In closed-condition responses, the model performed the counting task in earnest ("there are 12 pairs of ribs visible; I started from the topmost rib...") and simply did not mention the object; in open-condition responses, the same model on the same image flagged "an artificial object" overlaid on the scan. The open-condition response establishes that the object was reportable by that model on that item; the counting-condition omission establishes that the task set suppressed its report.

## Study 3: the gap is a threshold on output scope, not a monotonic load gradient

To probe the trigger of suppression, we varied the number of nodule-directed sub-questions from zero (open analysis) to three on the medical scenarios (two-judge $\kappa = 0.98$). The relationship was not a smooth dose-response (Figure 8). A single sub-question sufficed to switch suppression almost fully on: critical-finding report fell from 1.00 at zero questions to near zero at one. Adding further questions did not monotonically worsen suppression; the Anthropic models partially recovered (haiku 0.00 to 0.88 across one to three questions) while OpenAI models stayed at zero throughout. Response-length analysis clarified the pattern: the single narrow question produced the shortest outputs (mean 189 characters, rising to 435 and 1,046 as further questions were added, against 1,770 in the open condition) and the strongest suppression, whereas the longer outputs re-admitted the critical finding, sometimes as an explicit alert. The pattern therefore implicates output scope, the breadth of what the model commits to reporting, rather than question count alone: a narrow task compresses output onto the target and crowds out the unrequested signal. OpenAI models maintain this compression regardless of output length; Anthropic models relax it as output widens.

## Cross-cutting patterns

Combining the studies yields a map of when task-conditioned AI systems omit otherwise reportable signals. Suppression under an exclusive or attention-consuming instruction was observed across models and modalities; what varied was degree. In language, the focused instruction produced model- and salience-dependent suppression, while the strict instruction produced complete suppression everywhere. In vision, suppression was present in every model but strongest in the sampled OpenAI models: gpt-4o and gpt-4o-mini reported the unexpected object in 0.00 of counting-task items, whereas the sampled Anthropic models showed more partial suppression. Across the sampled systems, model family was a stronger correlate of the gap than model scale, alongside task load and signal salience (Figure 4). An apparent modality-by-provider reversal in small-N pilots did not survive scaling and two-judge adjudication. This heterogeneity is not a weakness of the construct but its resolution: the highest-risk setting is a high-load narrow task over a low-salience but safety-critical signal handled by a model family that more strongly compresses output to the requested target.

## The gap is genuine omission, not confabulation, and decouples from specified-task performance

Two controls establish that the gap is a genuine suppression of a detectable signal rather than an artifact of over-reporting or of an inability to do the requested task. First, a signal-absent control tests whether open-condition reports reflect true detection. We presented the valid base images with no composited object (11 of the 12 source identities: one source file was an invalid download artifact that errored in every query, and the multi-panel PNG source was excluded by the image-selection logic; Supplementary Note 2) to the four vision models under both instructions and asked judges whether each response nonetheless claimed an artificial, non-anatomical object (real medical hardware excluded). Under the counting task, the false-alarm rate was 0.00 in every model, so the task-conditioned omission observed in Study 2 is not accompanied by fabrication. Under the open instruction, false alarms were low in the Anthropic models (0.00 to 0.18) and higher in gpt-4o (0.36), indicating a yes-bias in the most permissive condition for that model. Combining object-present hits matched to the same 11 source identities (44 object-present trials per model) with object-absent false alarms yields positive signal-detection sensitivity in every model ($d' = 1.38$ to $2.32$; Table 3), so the open-condition report reflects genuine discrimination and the within-item gap survives correction for response bias. The gap is therefore a true task-conditioned suppression of a signal the model can detect, not a byproduct of open-condition over-reporting.

Second, the suppression coexists with successful performance of the requested task. Re-adjudicating the radiology focused-condition responses for whether the model completed the requested nodule work-up (two judges, raw agreement 0.93), the models produced the work-up in 91% of items overall while omitting the co-present critical finding: the two frontier OpenAI models completed the requested work-up in 0.88 (gpt-4o) and 0.91 (gpt-4o-mini) of items while omitting the co-present life-threatening finding in 0.80 and 0.81, a within-item decoupling of specified-task competence from unspecified-hazard reporting in 0.69 to 0.75 of items. A model can thus discharge its assigned task almost perfectly while remaining silent on a co-present hazard, the operational content of the claim that specified-task performance and unspecified-hazard reporting diverge.

## Cross-vendor and flagship validation

To test whether the effect is an artifact of one vendor or of model scale, we ran the medical text study on three flagship models from three vendors (Opus 4.8, GPT-5, Gemini 2.5 Pro; two-judge $\kappa = 0.97$). The provider pattern held and sharpened. GPT-5, a reasoning model, showed complete suppression in valid responses: focus and strict report rates were 0.00 versus open 1.00. The strict-condition comparison was significant on 21 paired items after reasoning-budget attrition (exact McNemar $p = 9.5e-7$), and the focused-condition comparison was significant on the 9 paired items remaining after attrition ($p = 3.9e-3$). On the strict comparison, this refutes, in the sampled setting, the hypothesis that inference-time reasoning alone resolves the gap; the focused-condition result on the 9 surviving pairs is consistent with the same conclusion. Gemini 2.5 Pro, a third-vendor model, suppressed nearly as strongly (0.00 to 0.09; exact $p = 1.2e-7$ strict, $9.5e-7$ focused). Opus 4.8, by contrast, was robust, reporting the co-present finding in 0.83 to 0.96 of task-conditioned items and 1.00 of open items. Inspection showed the behavioral contrast directly: GPT-5 produced a clean nodule-only work-up and omitted the co-present pneumothorax, whereas Opus 4.8 prefaced its report with “a critical finding that I can’t ethically ignore despite the nodule-only request” and flagged it. The determinant is therefore best described as a model-family-associated safety-reporting disposition rather than simple scale: within Anthropic, suppression did not increase with capability, while OpenAI models suppressed strongly at every sampled scale, including GPT-5. A testable training-side account is that families weight instruction-following and safety-reporting objectives differently during alignment; an override that survives an adversarial suppression instruction at 0.96 is consistent with the signature a prompt-robust safety-reporting objective would leave, and the hypothesis is directly examinable by vendors with access to training recipes. GPT-5’s reasoning-budget attrition was handled by excluding empty responses from paired analyses; suppression in valid responses was unambiguous.

## A System-1-style task capture without reliable intrinsic oversight

In dual-process terms, human inattentional blindness can be read as a failure of deliberate monitoring under a primary attentional set: the process that would catch the unexpected is consumed by the task (53). We asked whether conditioned models exhibit any analogous oversight behavior, using three probes on the medical scenarios. First, inference-time reasoning did not supply such oversight in the sampled reasoning model. GPT-5 showed complete suppression in valid responses under both focused and strict instructions, indicating that a reasoning-capable inference pass remained inside the assigned task rather than acting as a boundary monitor. Second, one sampled model family exhibited a robust safety-reporting override, most clearly in Opus 4.8. Opus 4.8 reported the co-present critical finding in 0.96 of items even under an adversarial-suppression instruction that explicitly forbade mentioning any other finding or safety caveat and scored such mentions as failures, matching its 0.96 under standard focus and approaching its 1.00 open rate (24 scenarios, inter-judge agreement 0.97). Strengthening the suppression instruction therefore did not break the override, suggesting a robust model-specific safety-reporting disposition rather than a fragile prompt reflex. No comparable override appeared in the other sampled models. Third, oversight could be supplied externally. Routing each narrow report from a task model (gpt-4o-mini, which reported the co-present finding in 0.00 of items on its own) to an independent open-ended critic that re-examined the same input for omissions raised the pipeline's report rate to 1.00 (McNemar $p = 1.2e-7$); the critic recovered every omission the single pass had dropped. This recovery is not the trivial consequence of a critic that flags something on every case: on 24 matched hazard-absent scenarios, in which the imaging contained no co-present critical finding beyond the nodule, the same claude-sonnet-4-6 critic used in the recovery pipeline raised a false alert in 0.12 of cases (specificity 0.88 under the both-judge consensus used throughout; 0.83 under a conservative either-judge rule), and none of the three false alerts fabricated a co-present acute finding: one disputed the categorization and follow-up recommended for the specified nodule, and two flagged omitted negative or stability statements. The scaffold therefore combines high recovery with high specificity, the two properties a usable safety layer requires.

Together, these probes locate the gap as System-1-style task capture rather than as a failure eliminated by scale or inference-time reasoning alone (Figure 7). In the sampled systems, reliable oversight did not appear as an architecture-general behavior. Where monitoring appeared, it took one of two forms: a model-specific safety-reporting disposition or an explicit second process. The scaffold that restored reporting was the open-ended review run as a standing parallel analysis. This returns to the structural cost identified by the construct: if the architecture does not reliably supply boundary monitoring, deployment must implement and budget it as an explicit process.

## Discussion

The Inattentional Gap reframes what AI safety evaluation measures. A model that scores well on a targeted benchmark is being measured primarily on D_specified; the events that produce harm often live in D_unspecified, the region task-conditioning can suppress from report. Our results show this suppression to be strong, statistically robust, and present in sampled frontier models, and they measure the decoupling directly: the frontier models completed the requested task in roughly nine of ten cases while omitting the co-present critical finding in four of five, so competence on the specified target and coverage of the unspecified hazard come apart within the same item. This decoupling compounds documented technical and epistemic shortcomings of current safety benchmarks (54) and aligns with the emerging community effort to redefine how AI is evaluated (55). The decoupling is sharpest where it is most consequential: a diagnostic AI told to assess one finding omits a life-threatening second finding it can otherwise describe; a vision-language model performing a primary task fails to report an unexpected object in plain view.

The phenomenon mirrors human inattentional blindness at the level of behavior, but the mechanism is unlikely to be identical. Human inattentional blindness is classically explained by a capacity-limited attentional bottleneck; transformers do not instantiate that same human bottleneck. The behavioral convergence with a different likely mechanism is therefore informative, and consistent with prior arguments that performance parity need not imply architectural equivalence (7). The convergence is quantitative as well as qualitative: roughly half of naive human observers fail to report the unexpected object (1), 83% of expert radiologists missed a composited gorilla (2), and the sampled models omitted the unexpected object in 46 to 100% of counting-task items in the star condition. Our results locate the machine-side effect not in a perceptual-capacity deficit, since the open condition recovers the signal, but in the conditioning of report on the task instruction (Figure 5). The family-dependence of the effect and the absence of a monotonic relationship with model scale suggest a model-family-associated reporting disposition rather than a scaling-limited capability boundary.

The clinical and operational implications are concrete. In radiology, suppression of co-present findings is the machine analogue of satisfaction of search and incidental-finding omission, long documented in human readers, where it reflects perceptual-cognitive rather than acuity limits (56, 57) and extends beyond radiology to real-time clinical monitoring (58). Here, we show that a comparable omission can be induced in AI by the framing of the request. In autonomous driving, the same structure predicts that a perception stack optimized for specified hazards may underreport out-of-distribution ones, the failure mode that the driving anomaly-detection literature (59) and dedicated anomaly-segmentation benchmarks (60) were built to expose. In both settings, the corrective is not simply a better detector of D_specified, but an evaluation and deployment regime that explicitly measures and protects D_unspecified.

This work also speaks to the question that motivated it: whether AI exhibits a human-like form of inattentional blindness. The evidence supports a behavioral analogue in the sampled systems: the behavioral parity is real, but the probes above indicate that the underlying mechanism differs. Humans miss critical signals when a capacity bottleneck exhausts deliberate monitoring (53); the sampled models omit them in a pattern most consistent with task-conditioning closing the reporting frame, and reliable boundary monitoring is not supplied by scale or inference-time reasoning alone. The productive reading is neither that AI cognition is human-like nor that it is alien, but that task-conditioning imposes a reporting closure that is structurally useful for compliance while hazardous for safety-critical deployment. Its failure modes can therefore be measured, mapped, and engineered against.

That the omission faithfully obeys the posed task does not soften the safety implication; it sharpens it. The more competently a model executes a narrow task, the more completely a mis-posed task may be obeyed, and the life-threatening finding goes unreported precisely because the system did what it was told. This is the machine form of the bias of the question posed (10). Safety is therefore not secured by making the model follow the narrow task more narrowly, nor by assuming an infeasible universal analyzer that reports everything on every input. It requires an evaluation and deployment regime, or an explicit dual-process scaffold, that measures and protects the unrequested region.

Our probes indicate that both components are implementable now. The external-critic scaffold, a second model re-examining each narrow report under an open instruction, restored every omitted finding in our medical scenarios at the cost of one additional inference pass. The within-item open/task contrast used throughout this paper is itself a deployable audit: any task-scoped system can be scored on the signals it fails to report relative to its own unconstrained reference on the same inputs. We term this reporting-complete evaluation, the requirement that a safety evaluation measure not only detection of the specified target but report of co-present unspecified hazards, and propose it as an admission criterion for task-scoped AI in safety-critical settings. Operationally, a certification set could require a task-scoped system to report co-present unspecified hazards at no less than a fixed fraction of its own open-condition reportability on the same items; at the measured critic specificity of 0.88, one additional inference pass per narrow report bought full recovery in our probe at a false-alert cost of 0.12, a trade that scales linearly with screening volume and should be budgeted against the alert-fatigue dynamics documented in the automation-bias literature.

Several limitations bound these claims. The studies remain moderate in scale (100 textual scenarios and 48 visual scenarios) and should be replicated at field scale. Adjudication used two independent language-model judges with high agreement (Cohen $\kappa = 0.90$ for text and 0.92 for vision); future work should add blinded human and domain-expert adjudication, especially for clinically realistic radiology tasks. The visual stimuli used composited non-anatomical objects rather than naturally occurring pathological findings, which isolates the attentional-set mechanism at the cost of clinical realism; the signal-absent control and signal-detection analysis (Table 3) establish that the open-condition reports reflect genuine discrimination ($d' = 1.38$ to 2.32) rather than response bias, but a complementary study using real co-present pathology on radiologist-adjudicated multi-finding datasets remains an important next step.

A pilot in this direction, run on real chest radiographs with no composited object and using each model's own open-condition description as the reportability reference, qualitatively reproduced the suppression on genuine radiographic content. For example, a model that reported cardiomegaly when asked openly omitted it under a rib-counting task. However, reliable quantification was limited by judge disagreement on open-ended clinical findings and by vision refusals on real radiographs. Radiologist-adjudicated multi-finding datasets would provide a stronger clinical test of the construct.

Open-condition report of the visual object was below ceiling for subtle stimuli, reflecting genuine low-salience difficulty. The within-item gap measure is robust to this limitation because suppression is estimated only relative to each model's own open-condition reportability. Two OpenAI models initially refused diagnostic framings until a synthetic-stimulus system message was supplied; this refusal behavior is itself deployment-relevant and is reported as part of the evaluation context.

The dual-process probes used 24 medical scenarios each and should be treated as demonstrations rather than full benchmarks. Scaling the external-critic scaffold across models, modalities, and domains, and testing whether the robust safety-reporting override observed in one model generalizes beyond these items, are immediate next steps. Finally, the construct is demonstrated behaviorally but not yet fully causally decomposed. Isolating the separate contributions of task load, signal salience, output scope, and model-family-specific reporting dispositions remains for future work.

## Methods

**Models.** Studies 1 to 3 used claude-haiku-4-5, gpt-4o-mini, gpt-4o, and claude-sonnet-4-6; the cross-vendor validation additionally used three flagship models, claude-opus-4-8, gpt-5 (a reasoning model), and gemini-2.5-pro. Models were queried at temperature 0 via their respective APIs, with temperature omitted for claude-opus-4-8 (which does not accept it) and for the gpt-5 reasoning model (whose output budget is set by max_completion_tokens). All models were queried in June 2026 against each vendor's production API; the complete per-call configuration (exact model identifiers, maximum output tokens, image-detail setting, and retry and refusal handling) is provided in the deposited code.

**Study 1 stimuli.** 100 procedurally composed scenarios (64 radiology: 8 nodule descriptions × 8 critical findings; 36 driving: 6 lead-vehicle scenes × 6 hazards). Three instructions per scenario (focused, strict, open). In the driving set the instructions were worded for the domain: focused ("report the lead vehicle's status"), strict exclusive ("report only the lead vehicle; do not mention anything else"), and open ("report all hazards"); each scenario paired the designated target with one co-present, unrequested safety-critical hazard.

**Study 2 stimuli.** 12 publicly available thoracic images (top and bottom text bands redacted to remove identifiers), each composited with a subtle translucent disc or yellow star at 4% image width, alpha ~0.59, at left or right lung field, yielding 48 images (12 scenes × 2 objects × 2 sides). A post hoc integrity audit of the frozen set identified two pairs of inadvertently duplicated scenes (10 unique) and two non-radiograph images (a labeled anatomical illustration and a photograph of radiographic equipment); the frozen stimuli are archived with SHA-256 manifests, and sensitivity analyses excluding the non-radiograph scenes and collapsing duplicates are reported in Supplementary Note 2. Two instructions per image (closed rib-counting, open description) under a synthetic-stimulus system message.

**Study 3 stimuli (output scope).** A subset of 24 radiology scenarios (4 nodule descriptions × 6 critical findings) was presented under four task-load levels defined by the number of nodule-directed sub-questions appended to a neutral report request (0 = open analysis; 1, 2, 3 cumulative). The number of sub-questions, and the resulting output length, were the manipulated variables; the dependent measure was report of the co-present critical finding.

**Cross-vendor (flagship) stimuli.** The same 24 radiology scenarios were presented to the three flagship models under the focused, strict, and open instructions of Study 1.

**Dual-process probes.** Two further experiments used the 24 medical scenarios. To test whether a robust model's override is a stable disposition or a fragile reflex, claude-opus-4-8 was run under a standard focused instruction, an adversarial-suppression instruction that explicitly forbade mentioning any other finding or any safety or ethical caveat (scoring such mentions as task failure), and an open instruction. To test whether external oversight closes the gap, each focused report from a task model (gpt-4o-mini) was passed, together with the original imaging description, to an independent open-ended critic model (claude-sonnet-4-6) instructed to flag any clinically critical finding omitted from the report; the pipeline counted as reporting the finding if either the task report or the critic flagged it, judged separately to avoid truncation of the critic text.

**Adjudication.** Each response was scored for report of the critical signal/object by two independent language-model judges (gpt-4o-mini and claude-haiku-4-5), each returning a binary decision; we report the both-agree consensus and inter-rater Cohen κ (0.895 Study 1 text, 0.918 Study 2 vision, 0.984 Study 3, 0.969 cross-vendor). Refusals and errors were excluded; the author's spot-checks validated judge decisions against response text. Because one judge (claude-haiku-4-5) is also a subject model, we re-scored every haiku-subject response using the non-haiku judge alone as a circularity check: all headline rates were unchanged (strict 0.00, radiology focused 0.53, star counting 0.50), and no rate moved by more than 0.05, ruling out self-adjudication as a driver of the results. Judges evaluated the first 1,500 characters of each stored response in Studies 1 to 3 and the flagship study (1,800 for the override probe; full responses for the specified-task re-run), a window sized to the response formats the task instructions elicit. The specified-task full-response re-run, conducted three weeks later, therefore doubles as a temporal robustness re-run under full-response adjudication and current production models; its focused-condition report rates for the OpenAI models (0.19 to 0.20, against the Study 1 open-condition reference of approximately 1.00) preserve the suppression pattern. In the dual-process probes, critical-finding report is near-constant within a condition, so Cohen's κ is degenerate (the prevalence paradox in which κ collapses toward zero despite high agreement); for these conditions we therefore report raw inter-judge agreement (0.97 for the override probe, 0.88 for the single-pass baseline) and, for the override probe, Gwet's AC1 (0.97), which is robust to skewed prevalence. Throughout, the n reported for a McNemar comparison denotes paired items in which both the task and the open condition returned a valid (non-refusal, non-error) response; for GPT-5 this paired n is 21 (strict) and 9 (focus) after reasoning-budget attrition, and is therefore smaller than the per-condition valid-response counts.

**Statistics.** Within-item paired comparison of closed versus open conditions by exact McNemar test (two-tailed, alpha = 0.05); report rates and condition gaps reported per model, object, and domain. Report rates carry 95% bootstrap confidence intervals (10,000 resamples; for example, radiology focused-condition rates were 0.08 for gpt-4o, item-level CI 0.02 to 0.14 and template-cluster CI 0.00 to 0.23, and 0.53 for haiku, item-level CI 0.41 to 0.66 and template-cluster CI 0.33 to 0.73, in every case separated from the open-condition rates). The more conservative template-cluster intervals are reproduced by the included cluster-bootstrap script, and the item-level intervals by the included item-bootstrap script. The thirteen primary McNemar comparisons (Study 1 strict and focused by model, Study 2 star by model, and the dual-process critic contrast) all survive Holm-Bonferroni correction across the family (maximum Holm-adjusted $p = 0.0156$). For Study 3, report rate is given per dose level, with the partial recovery at higher doses analyzed against response length. Because items were procedurally composed from a small set of templates, report rates were re-estimated with a cluster bootstrap resampling the nodule and critical-finding templates; the focused-condition gaps and the universal strict-condition suppression (report rate 0.00 across all eight nodule and all eight critical-finding templates) were robust to this clustering. The dual-process probes were also repeated across independent runs to check stability against temperature-0 non-determinism: the external-critic scaffold reproduced exactly in every run (single-pass report 0.00, pipeline report 1.00), and the robust-model override remained high throughout (0.92 to 1.00).

**Signal-absent control and signal-detection analysis.** To test whether open-condition object reports reflect true detection rather than a yes-bias, the valid base images (11 of the 12 source identities; Supplementary Note 2) with no composited object were presented to the four vision models under the same closed counting and open description instructions and system message as Study 2. Two independent judges scored whether each response claimed an artificial or non-anatomical object (a star, disc, geometric shape, or foreign marker); real medical hardware (tubes, lines, clips) and genuine anatomy were explicitly excluded from counting as a false alarm, using the both-agree consensus. Per model, the open-condition hit rate (from Study 2 object-present trials) and object-absent false-alarm rate were combined into sensitivity, specificity, and loglinear-corrected d' (Hautus, adding 0.5 to each cell; Table 3).

**Specified-task accuracy.** To test whether suppression coexists with successful task performance, the 64 radiology focused-condition scenarios were re-queried capturing the full model response, and two independent judges scored each response both for completion of the requested nodule work-up (report of specific nodule characteristics: size, margins, density, or lobe) and for report of the co-present critical finding ($\kappa = 0.91$ for the critical-finding judgment; raw inter-judge agreement 0.93 for task completion, whose Cohen $\kappa$ is deflated by the 0.91 prevalence of completed work-ups). An item was counted as decoupled when the work-up was completed and the critical finding omitted.

**Critic specificity.** To test whether the external-critic recovery is specific rather than a critic that alerts on every case, the critic pipeline was run on 24 matched hazard-absent scenarios (4 nodule descriptions × 6 clean, non-critical filler statements) containing no co-present critical finding. The critic was the same claude-sonnet-4-6 model and prompt as in the recovery experiment; judges were gpt-4o-mini and claude-haiku-4-5, so the critic model did not adjudicate its own output. Two judges scored whether the critic asserted that a specific critical finding was present and omitted (a false alert); the both-agree false-alert rate was 0.12 (specificity 0.88; inter-judge $\kappa = 0.83$), or 0.17 under a conservative either-judge rule (specificity 0.83).

**Use of artificial intelligence tools.** Large language and vision-language models are the objects of study in this work and were queried programmatically as experimental subjects. Separately, generative AI assistants were used for code scaffolding and language editing during manuscript preparation. All figures and statistical results were generated from the deposited analysis scripts.

**Ethics.** This study analyzed the behavior of commercial AI models; no human participants were recruited and no patient records were accessed. Visual stimuli were derived from publicly posted thoracic images with identifier bands masked in every distributed stimulus; one archived source image later found to retain identifiers was restricted in the public deposit, and stimulus images are now available for confidential editorial inspection and, where redistribution rights permit, research verification (Supplementary Note 2). A subsequent audit found that three archived model-response excerpts transcribed the identifier band of that source image; the transcribed identifiers are redacted in the current public deposit, earlier deposit versions carrying them are access-restricted, and the redaction is verified by a full-package scan.

## Data availability

The per-item adjudication records, including response excerpts for every trial and full response text for the specified-task re-run, that support the findings of this study are archived under CC-BY 4.0 at Zenodo (https://doi.org/10.5281/zenodo.20826823; this DOI resolves to the latest version, and the version archived at submission is v2.1, https://doi.org/10.5281/zenodo.21754672). Source images were collected from publicly posted online material; because per-file redistribution rights could not be documented, the stimulus images (source and composited) are not redistributed in the public deposit and are available for confidential editorial inspection and, where redistribution rights permit, research verification via the corresponding author, with one disclosed exception: the source identity shown in Figure 3 was verified post hoc as byte-identical to a CC0 Wikimedia Commons original, and a rights-verified reconstruction of that stimulus appears in Figure 3. The deposit anchors the frozen stimulus set with SHA-256 manifests and archives the compositing and redaction script, a source-provenance statement, and the stimulus-integrity audit (Supplementary Note 2); all per-item adjudication records are public, with three response excerpts redacted where they transcribed identifiers from a source image, so every reported statistic reproduces without the images. Because commercial production models may change over time, the deposited per-item adjudication records constitute the authoritative evidence base for the reported analyses. All other study data are included in the article and/or Supplementary Information.

## Code availability

Scenario generators, image-composition and stimulus-generation code, model-query scripts, adjudication scripts including the verbatim judge prompts, analysis and robustness scripts, and figure-generation scripts are available in the same Zenodo archive (https://doi.org/10.5281/zenodo.20826823; submission snapshot v2.1). API rerun scripts are provided for reproducibility and extension.

## Acknowledgments

This research received no external funding.

**Author contributions.** K.S.S. conceived the study, designed the experiments, implemented the code, conducted the analyses, wrote the manuscript, and approved the final version.

**Competing interests.** The author declares no competing interests.

## Figures

**Figure 1. The Inattentional Gap (schematic).** Task-conditioning focuses evaluation on the specified target, where report on the specified target appears high, while report of co-present unspecified safety-critical signals is suppressed, decoupling specified-target performance from reporting completeness.

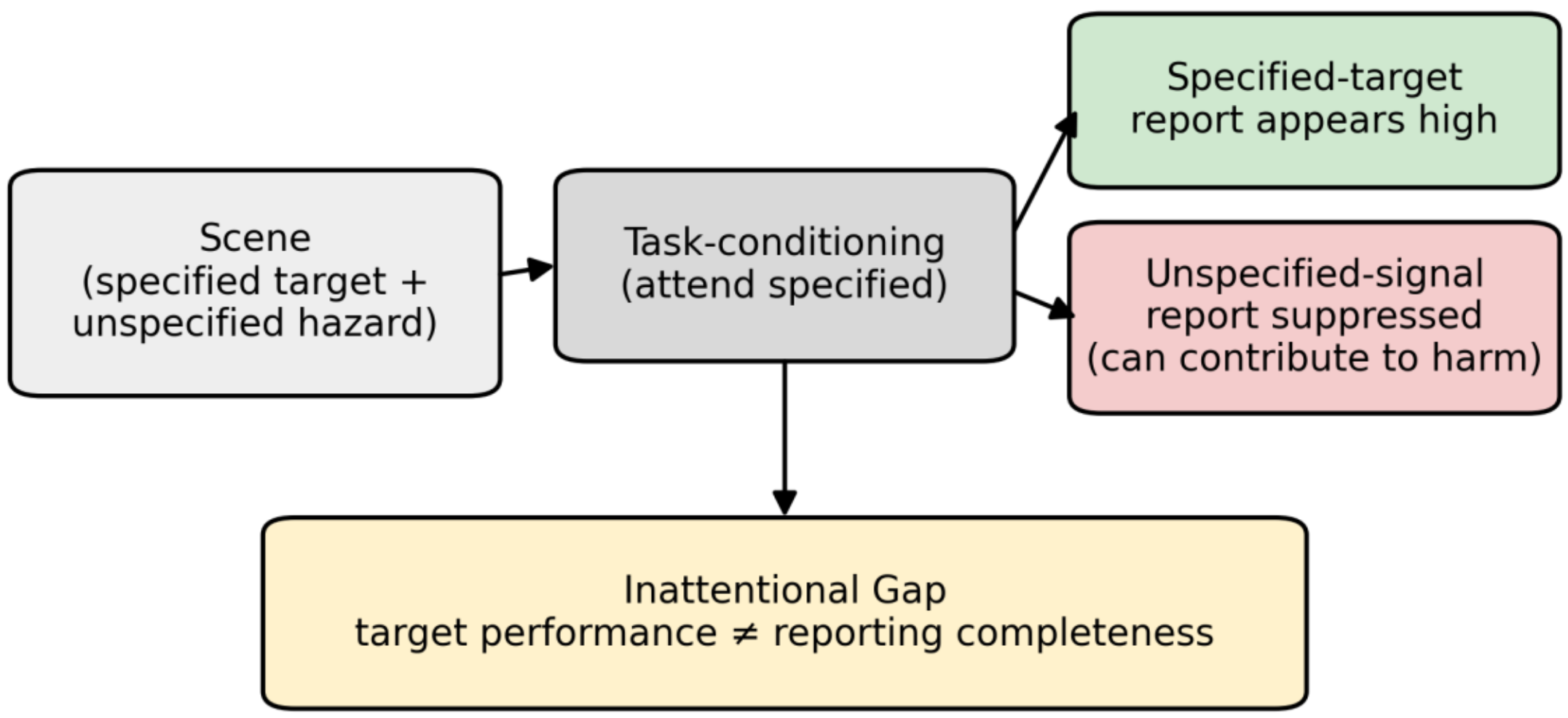

*Figure 1*

**Figure 2. Study 1 (text): critical-signal report rate by model and condition.** Low values indicate stronger inattentional suppression. Open-condition columns are near ceiling, establishing the reportability reference; task-conditioned columns show suppression, strongest under strict instructions and in the radiology focused condition.

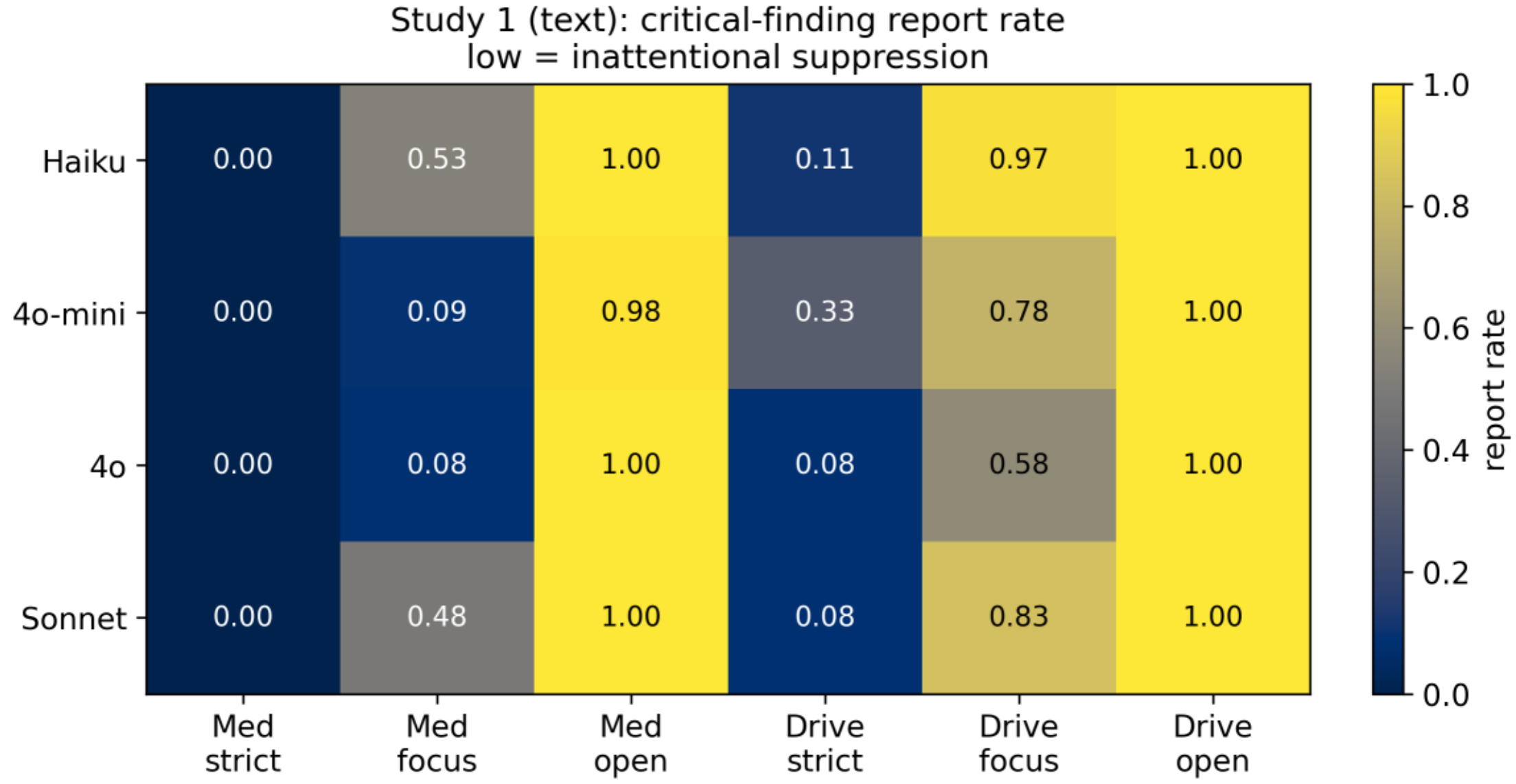

*Figure 2*

**Figure 3. Study 2 (vision): stimulus and star-object report rate.** Panel a is a rights-verified reconstruction of one frozen experimental stimulus identity: its source image was verified post hoc as byte-identical to a public-domain original (Mikael Häggström,

Wikimedia Commons, CC0), and the panel recomposites that source with the same parameters and redaction geometry as the frozen stimulus, differing from the exact frozen file only in JPEG encoding. The exact frozen files and all remaining third-party-derived stimuli are withheld and hash-anchored in the deposit (Supplementary Note 2). Panel b shows star-object report rates under the rib-counting task versus the open-description condition; the disc condition is summarized in the text.

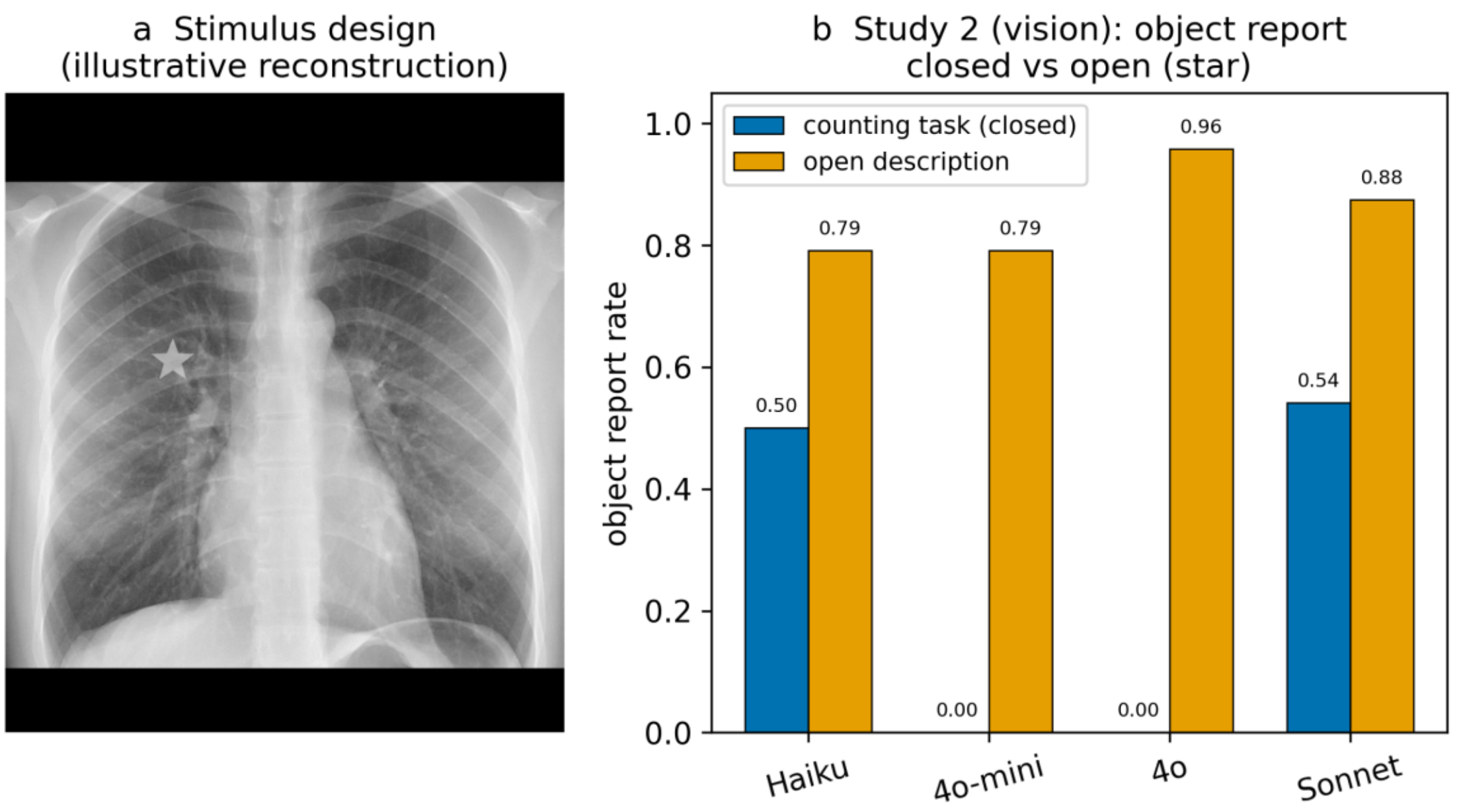

*Figure 3*

**Figure 4. Inattentional Gap by modality and model.** Higher values indicate stronger task-conditioned suppression. In the non-strict/task-loaded conditions, the sampled OpenAI models showed larger gaps than the sampled Anthropic models in both text and vision settings; the cross-vendor validation further tests whether this pattern tracks model family rather than model scale.

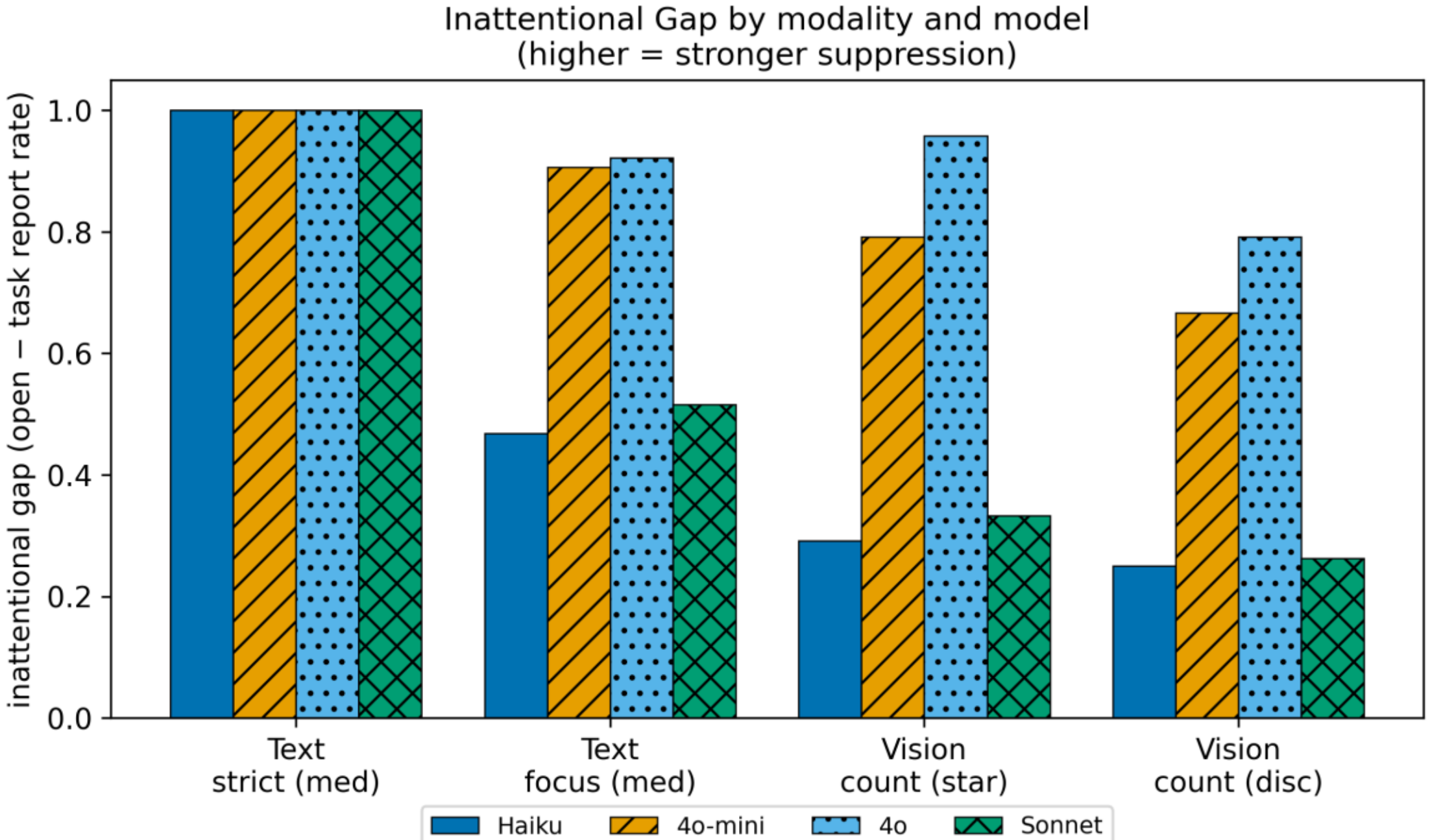

*Figure 4*

**Figure 5. Mechanism.** Under an open instruction, the model reports both the specified target and the co-present safety-critical signal. Under a task-conditioned instruction, the same input is compressed into the requested reporting frame, and the unrequested hazard is omitted. The gap is therefore located in task-conditioned reporting, not in failure to report the signal under open conditions.

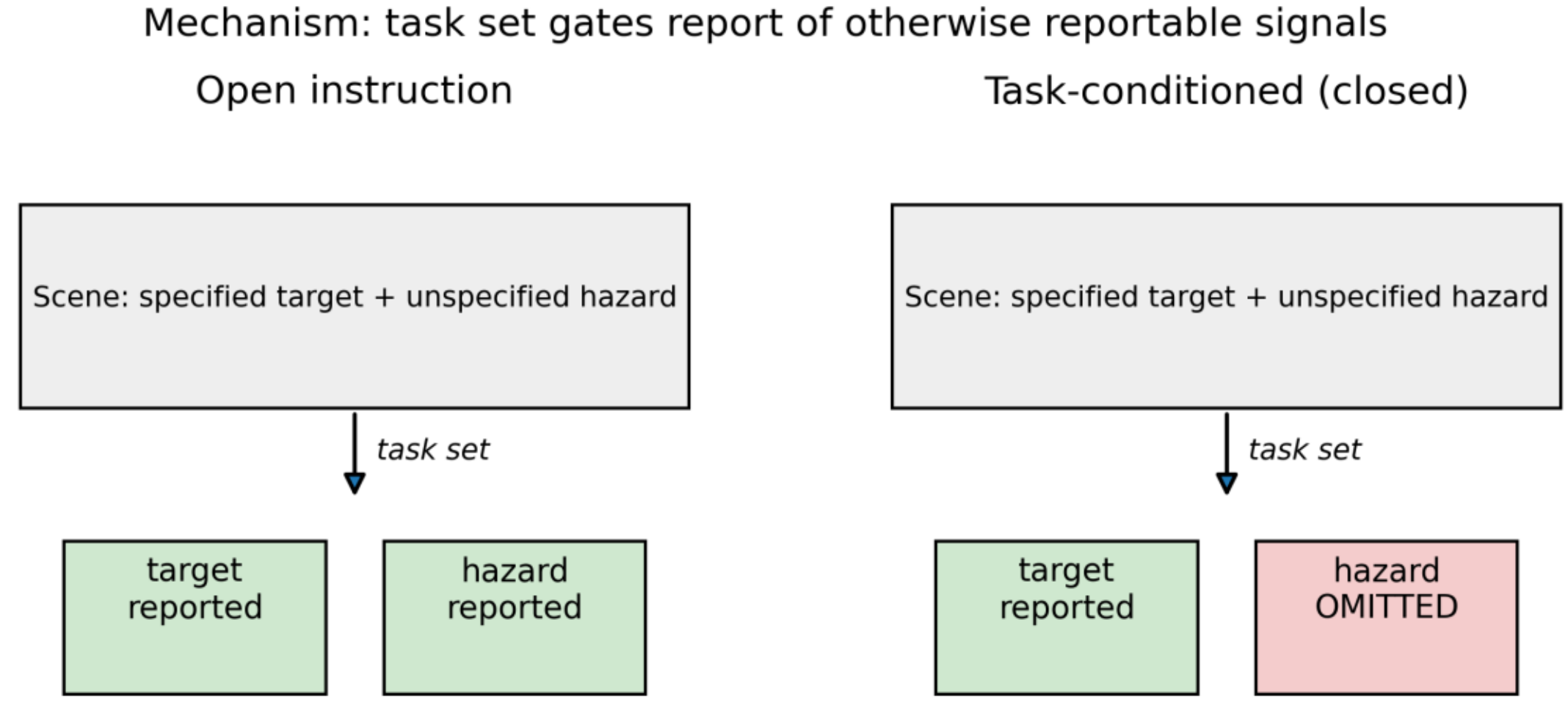

*Figure 5*

**Figure 6. Structured positioning map of the literature.** We coded a 116-work core corpus across eleven axes, rendered as twelve thematic clusters, and separately added two recent nearest neighbors that appeared before submission (51, 52). The x-axis captures engagement with inattentional-blindness theory; the y-axis captures whether task-conditioning is used as the independent variable with an open/closed within-item reportability control. Human inattentional-blindness studies occupy the high-x, high-y region, whereas machine studies, including recent neighbors that invoke inattentional blindness or document compliance-driven omission (51, 52), stop short of the reportability-controlled, safety-critical corner of that cell in our coding. The Inattentional Gap addresses that underdeveloped machine-side cell by combining attentional-set manipulation, within-item reportability recovery, and safety-critical framing. Coordinates summarize the coding rationale reported in Supplementary Information and should be read as a structured positioning map rather than a quantitative meta-analysis.

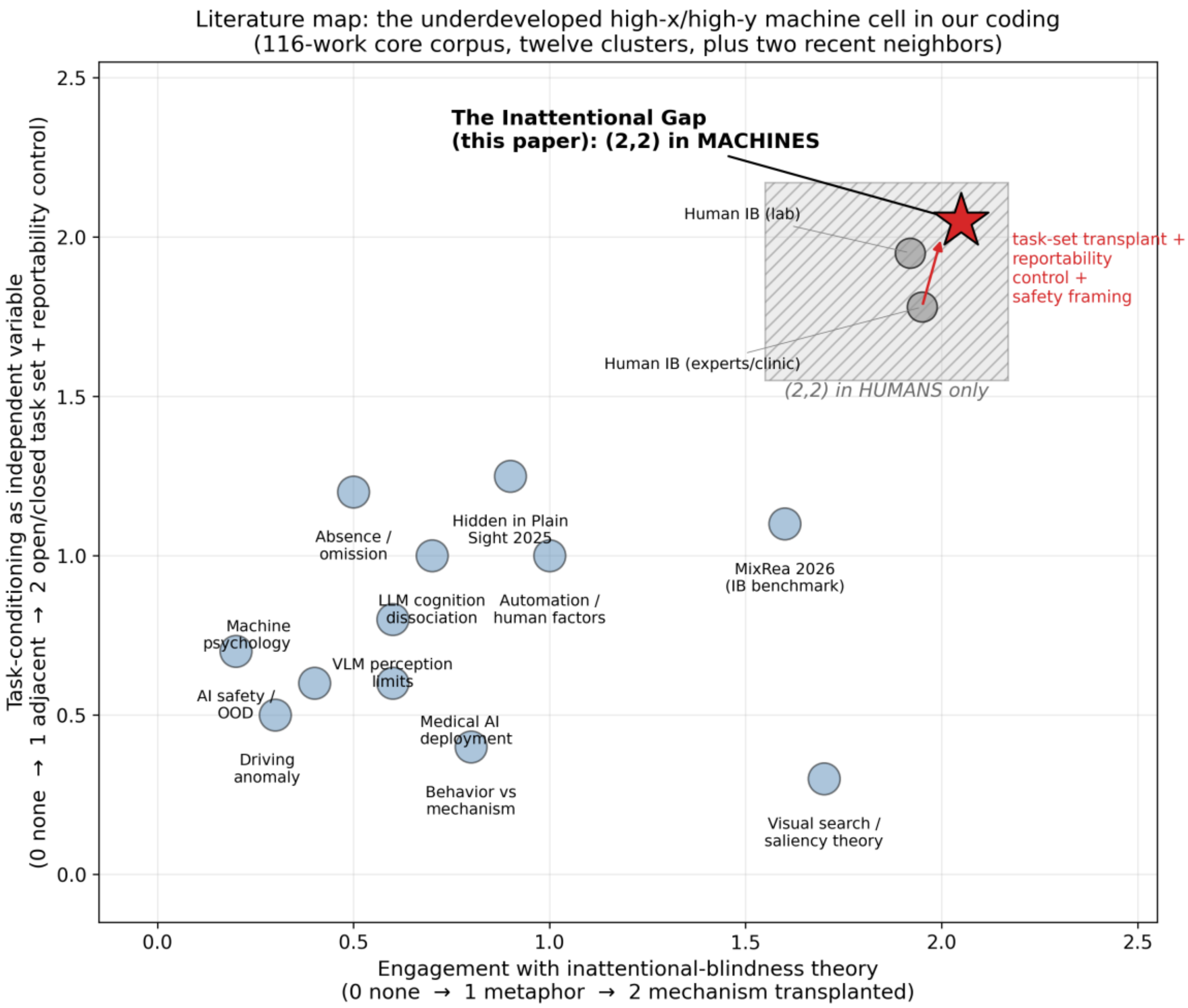

*Figure 6*

**Figure 7. Dual-process probes.** (a) In valid responses, the sampled reasoning model (GPT-5) showed complete suppression under both focused and strict task instructions, whereas Opus 4.8 reported the co-present critical finding even under an adversarial-

suppression instruction that explicitly discouraged reporting other findings or safety caveats, indicating a robust safety-reporting override rather than a fragile prompt reflex. (b) Routing a task model's narrow report to an independent open-ended critic raised the co-present-finding report rate from 0.00 to 1.00 (McNemar p = 1.2e-7), showing that oversight-like recovery can be supplied externally as a second process.

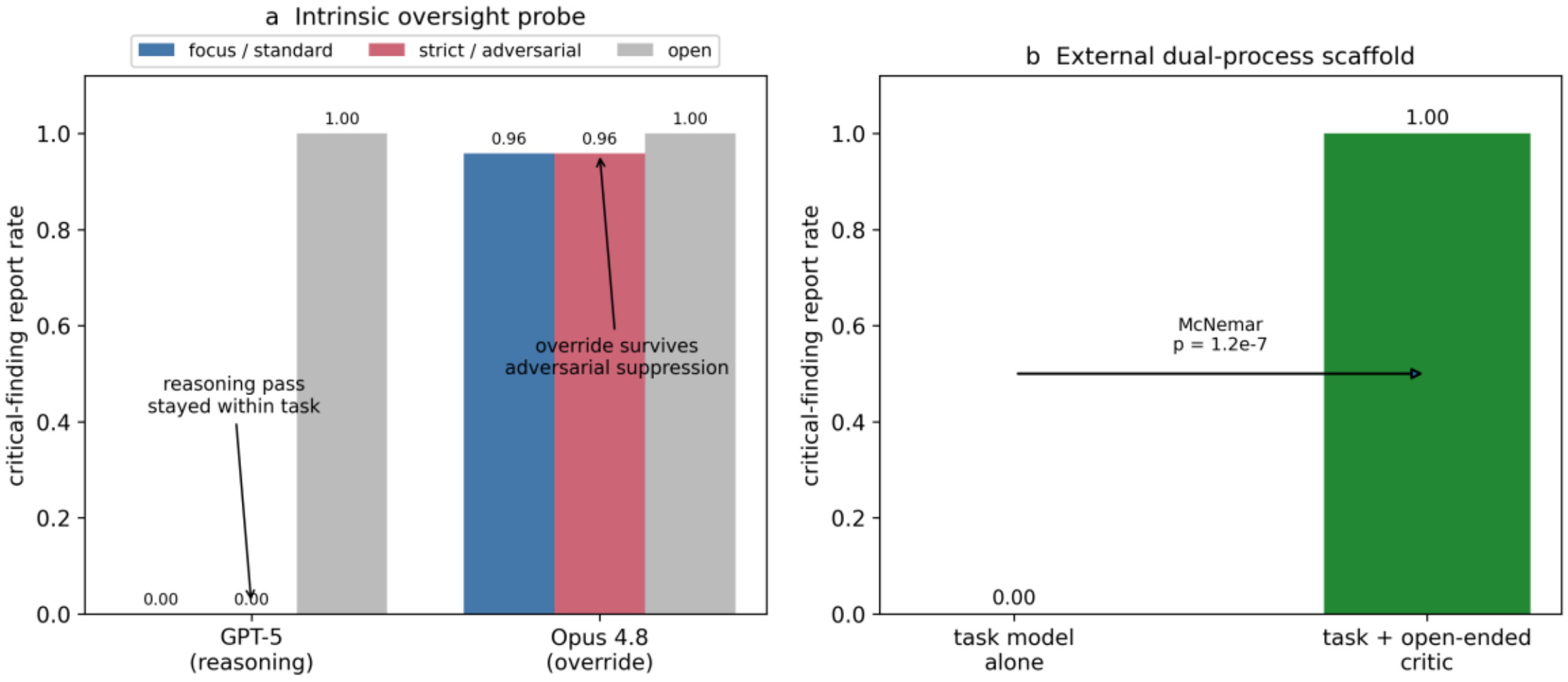

*Figure 7*

**Figure 8. Study 3: output scope, not monotonic load.** Critical-finding report rate (left axis, solid) collapses from 1.00 at zero sub-questions to near zero at one, then partially recovers for the Anthropic models as output scope widens with longer responses (mean output length, right axis, dashed); the OpenAI models stay suppressed. The pattern implicates the breadth of committed output, rather than the number of instructions alone, as the best-supported proximal behavioral account of suppression.

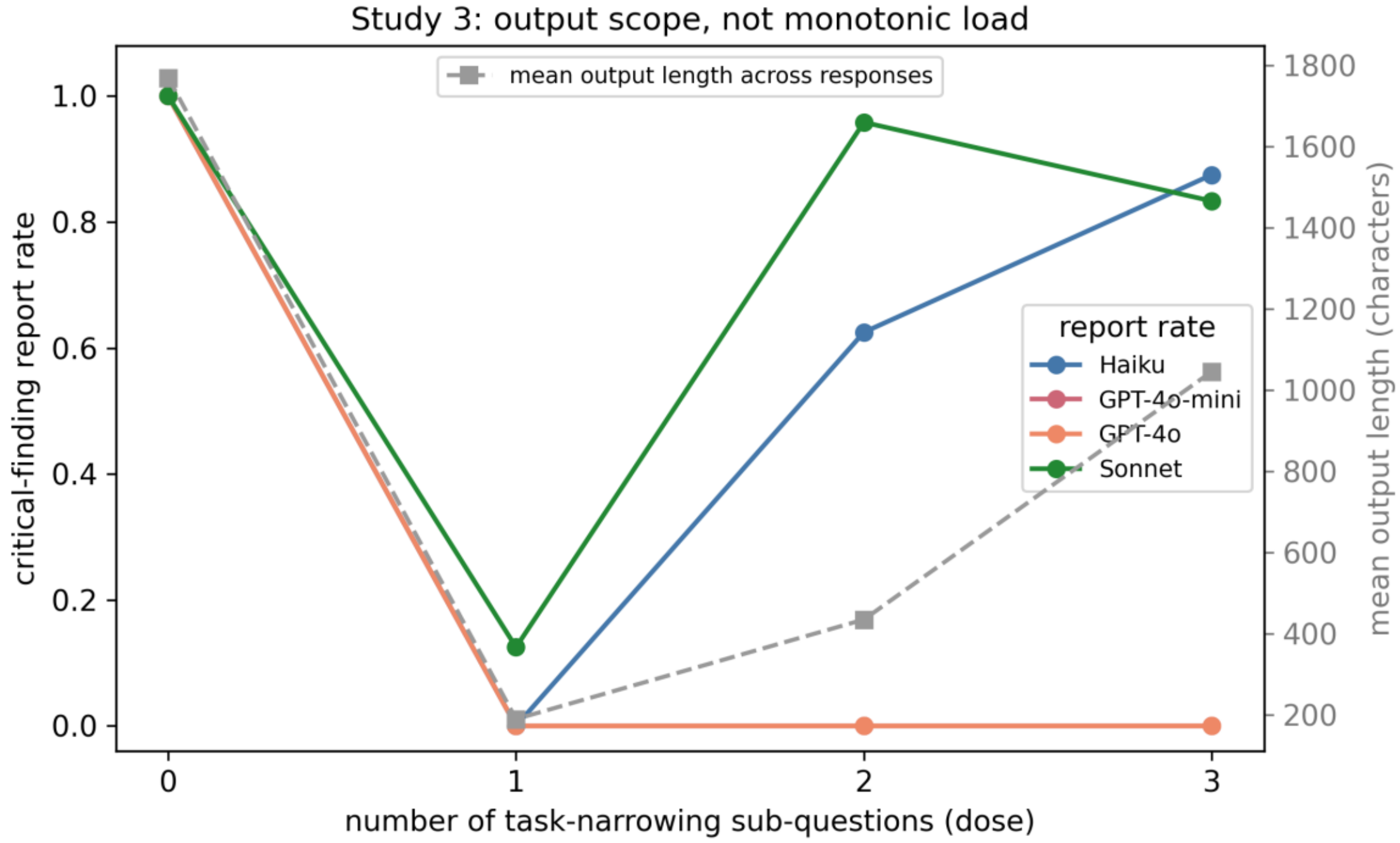

*Figure 8*

## References

1. Simons, D. J. & Chabris, C. F. Gorillas in our midst: Sustained inattentional blindness for dynamic events. Perception 28, 1059–1074 (1999).
2. Drew, T., Võ, M. L.-H. & Wolfe, J. M. The invisible gorilla strikes again: Sustained inattentional blindness in expert observers. Psychological Science 24, 1848–1853 (2013).
3. Mack, A. & Rock, I. Inattentional Blindness (MIT Press, 1998).
4. Binz, M. & Schulz, E. Using cognitive psychology to understand GPT-3. Proc. Natl. Acad. Sci. U.S.A. 120, e2218523120 (2023).
5. Hagendorff, T., Fabi, S. & Kosinski, M. Human-like intuitive behavior and reasoning biases emerged in large language models but disappeared in ChatGPT. Nature Computational Science 3, 833–838 (2023).
6. Dahou, Y. et al. Vision-language models can't see the obvious (SalBench). Proc. IEEE/CVF Int. Conf. Comput. Vis. (ICCV) (2025).
7. Bowers, J. S. et al. Deep problems with neural network models of human vision. Behavioral and Brain Sciences 46, e385 (2023).

8. Quilty-Dunn, J., Porot, N. & Mandelbaum, E. The best game in town: The re-emergence of the language-of-thought hypothesis across the cognitive sciences. Behavioral and Brain Sciences 46, e261 (2023).

9. Most, S. B., Scholl, B. J., Clifford, E. R. & Simons, D. J. What you see is what you set: Sustained inattentional blindness and the capture of awareness. Psychological Review 112, 217–242 (2005).

10. Zamir, E. The bias of the question posed: A diagnostic “invisible gorilla.” Diagnosis 1, 245–248 (2014).

11. Garg, R. K. et al. Inattentional blindness to DWI lesions in spontaneous intracerebral hemorrhage. Neurological Sciences 43, 4355–4361 (2022).

12. Hults, C. M. et al. Inattentional blindness in medicine. Cognitive Research: Principles and Implications 9, 18 (2024).

13. Neisser, U. & Becklen, R. Selective looking: Attending to visually specified events. Cognitive Psychology 7, 480–494 (1975).

14. Berbaum, K. S. et al. Satisfaction of search in diagnostic radiology. Invest. Radiol. 25, 133–140 (1990).

15. Berbaum, K. S., Franken, E. A., Caldwell, R. T. & Schartz, K. M. Gaze dwell times on acute trauma injuries missed because of satisfaction of search. Acad. Radiol. 8, 304–314 (2001).

16. Adamo, S. H., Gereke, B. J., Shomstein, S. & Schmidt, J. From “satisfaction of search” to “subsequent search misses”: A review of multiple-target search errors across radiology and cognitive science. Cogn. Res. Princ. Implic. 6, 59 (2021).

17. Williams, L. H. et al. The invisible breast cancer: Experience does not protect against inattentional blindness to clinically relevant findings in radiology. Psychonomic Bulletin & Review 28, 503–511 (2021).

18. Robson, S. G. & Tangen, J. M. The invisible 800-pound gorilla: Expertise can increase inattentional blindness. Cogn. Res. Princ. Implic. 8, 33 (2023).

19. Lake, B. M., Ullman, T. D., Tenenbaum, J. B. & Gershman, S. J. Building machines that learn and think like people. Behavioral and Brain Sciences 40, e253 (2017).

20. Hagendorff, T. et al. Machine psychology: Investigating emergent capabilities and behavior in large language models using psychological methods. Preprint at arXiv:2303.13988 (2023).

21. Lampinen, A. K. et al. Language models, like humans, show content effects on reasoning tasks. PNAS Nexus 3, pgae233 (2024).

22. Mahowald, K. et al. Dissociating language and thought in large language models. Trends in Cognitive Sciences 28, 517–540 (2024).

23. Yildirim, I. & Paul, L. A. From task structures to world models: What do LLMs know? Trends Cogn. Sci. 28, 404–415 (2024).

24. Webb, T., Holyoak, K. J. & Lu, H. Emergent analogical reasoning in large language models. Nat. Hum. Behav. 7, 1526–1541 (2023).

25. Block, N. Perceptual consciousness overflows cognitive access. Trends in Cognitive Sciences 15, 567–575 (2011).

26. Cohen, M. A., Dennett, D. C. & Kanwisher, N. What is the bandwidth of perceptual experience? Trends Cogn. Sci. 20, 324–335 (2016).

27. van Amsterdam, W. A. C., van Geloven, N., Krijthe, J. H., Ranganath, R. & Cinà, G. When accurate prediction models yield harmful self-fulfilling prophecies. Patterns 6, 101229 (2025).

28. Park, P. S., Goldstein, S., O'Gara, A., Chen, M. & Hendrycks, D. AI deception: A survey of examples, risks, and potential solutions. Patterns 5, 100988 (2024).

29. Sanchez, M. et al. AI-clinician collaboration via disagreement prediction: A decision pipeline and retrospective analysis of real-world radiologist-AI interactions. Cell Reports Medicine 4, 101207 (2023).

30. Rahmanzadehgervi, P., Bolton, L., Taesiri, M. R. & Nguyen, A. T. Vision language models are blind. Proc. Asian Conf. Comput. Vis. (ACCV) (2024).

31. Tong, S. et al. Eyes wide shut? Exploring the visual shortcomings of multimodal LLMs. Proc. IEEE/CVF Conf. Comput. Vis. Pattern Recognit. (CVPR) (2024).

32. Li, Y. et al. Evaluating object hallucination in large vision-language models. Proc. Conf. Empir. Methods Nat. Lang. Process. (EMNLP) (2023).

33. Fu, H. Y. et al. AbsenceBench: Language models can't tell what's missing. Preprint at arXiv:2506.11440 (2025).

34. Geirhos, R. et al. Shortcut learning in deep neural networks. Nat. Mach. Intell. 2, 665–673 (2020).

35. Amodei, D. et al. Concrete problems in AI safety. Preprint at arXiv:1606.06565 (2016).

36. Booth, S. et al. The perils of trial-and-error reward design: Misdesign through overfitting and invalid task specifications. Proc. AAAI Conf. Artif. Intell. 37, 5920–5929 (2023).

37. Oakden-Rayner, L., Dunnmon, J., Carneiro, G. & Ré, C. Hidden stratification causes clinically meaningful failures in medical imaging deep learning. Proc. ACM Conf. Health Inference Learn. (CHIL) 151–159 (2020).

38. Wu, E. et al. How medical AI devices are evaluated: Limitations and recommendations from an analysis of FDA approvals. Nat. Med. 27, 582–584 (2021).

39. Agrawal, M., Chen, I. Y., Gulamali, F. & Joshi, S. The evaluation illusion of large language models in medicine. npj Digit. Med. 8, 600 (2025).

40. Jin, Q. et al. Hidden flaws behind expert-level accuracy of multimodal GPT-4 vision in medicine. npj Digit. Med. 7, 190 (2024).

41. Draelos, R. L. et al. Large language models provide unsafe answers to patient-posed medical questions. npj Digit. Med. 9, 241 (2026).

42. Vaswani, A. et al. Attention is all you need. Adv. Neural Inf. Process. Syst. 30 (2017).

43. Evans, J. St. B. T. & Stanovich, K. E. Dual-process theories of higher cognition: Advancing the debate. Perspect. Psychol. Sci. 8, 223–241 (2013).

44. Itti, L. & Koch, C. A model of saliency-based visual attention for rapid scene analysis. IEEE Trans. Pattern Anal. Mach. Intell. 20, 1254–1259 (1998).

45. Bainbridge, L. Ironies of automation. Automatica 19, 775–779 (1983).

46. Parasuraman, R. & Riley, V. Humans and automation: Use, misuse, disuse, abuse. Human Factors 39, 230–253 (1997).

47. Parasuraman, R. & Manzey, D. H. Complacency and bias in human use of automation: An attentional integration. Hum. Factors 52, 381–410 (2010).

48. Dzindolet, M. T., Peterson, S. A., Pomranky, R. A., Pierce, L. G. & Beck, H. P. The role of trust in automation reliance. Int. J. Hum.-Comput. Stud. 58, 697–718 (2003).

49. Goddard, K., Roudsari, A. & Wyatt, J. C. Automation bias: A systematic review of frequency, effect mediators, and mitigators. J. Am. Med. Inform. Assoc. 19, 121–127 (2012).

50. Gaube, S. et al. Do as AI say: susceptibility in deployment of clinical decision-aids. npj Digit. Med. 4, 31 (2021).

51. Cai, Y. et al. MixRea: Benchmarking explicit-implicit reasoning in large language models. Preprint at arXiv:2605.20128 (2026).

52. Yan, Q. et al. Hidden in plain sight: Reasoning in underspecified and misspecified scenarios for multimodal LLMs. In Proc. 2025 Conference on Empirical Methods in Natural Language Processing 24715–24735 (Association for Computational Linguistics, 2025).

53. Kahneman, D. Thinking, Fast and Slow (Farrar, Straus and Giroux, 2011).

54. Yu, C., Engelmann, S., Cao, R., Ali, D. & Papakyriakopoulos, O. How should AI safety benchmarks benchmark safety? Preprint at arXiv:2601.23112 (2026).

55. Torres, I. & Evals-Consensus.AI Consortium. A call to join a collective effort on AI evaluation. Patterns 7, 101512 (2026).

56. Kundel, H. L., Nodine, C. F. & Carmody, D. Visual scanning, pattern recognition and decision-making in pulmonary nodule detection. Invest. Radiol. 13, 175–181 (1978).

57. Krupinski, E. A. Current perspectives in medical image perception. Atten. Percept. Psychophys. 72, 1205–1217 (2010).

58. de Cassai, A. et al. Inattentional blindness in anesthesiology: A gorilla is worth one thousand words. PLoS ONE 16, e0257508 (2021).

59. Bogdoll, D., Nitsche, M. & Zöllner, J. M. Anomaly detection in autonomous driving: A survey. Proc. IEEE/CVF Conf. Comput. Vis. Pattern Recognit. Workshops (CVPRW) (2022).

60. Chan, R. et al. SegmentMeIfYouCan: A benchmark for anomaly segmentation. Adv. Neural Inf. Process. Syst. Datasets Benchmarks Track (2021).

## Tables

**Table 1. Study design across the six primary experiments.** Three further controls (signal-absent, specified-task adjudication, critic specificity) are described in Methods. Every comparison is made within item; report of the co-present safety-critical signal is the dependent measure. In the open/closed studies, the open condition serves as the within-item reportability reference; in the dual-process β probe, the comparison is task-only versus task-plus-critic.

| Experiment | Modality | Stimuli | Conditions | Domain | Model configuration |
|---|---|---|---|---|---|
| Study 1 | Language | 100 (64 radiology; 36 driving, Supplementary Note 1) | focused, strict, open | radiology (driving generalization in SI) | 4 models |
| Study 2 | Vision | 48 (12 stimulus identities [10 | counting, open | thoracic imaging | 4 models |

| Experiment | Modality | Stimuli | Conditions | Domain | Model configuration |
|---|---|---|---|---|---|
| | | unique scenes] × 2 objects × 2 sides) | | | |
| Study 3 (output scope) | Language | 24 | 0, 1, 2, 3 sub-questions | radiology | 4 models |
| Cross-vendor | Language | 24 | focused, strict, open | radiology | 3 flagship models |
| Dual-process α | Language | 24 | standard, adversarial-suppression, open | radiology | Opus 4.8 |
| Dual-process β | Language | 24 | task-only, task + critic | radiology | task model + critic |

**Table 2. Critical-signal report rates by condition (two-judge consensus).** Low values under a task instruction indicate inattentional suppression; in open/closed studies, the open condition is the within-item reportability reference. Refusals and errors were excluded before adjudication. For GPT-5, some responses were empty because reasoning consumed the output budget and were excluded from paired analyses. Open-column rates use all valid open-condition responses; n denotes the paired items entering each McNemar comparison. Ninety-five percent bootstrap confidence intervals for all rates are reported in Methods and can be recomputed from the deposited per-item files.

| Experiment | Model | Task-condition report rate | Open | Key statistic |
|---|---|---|---|---|
| Study 1 (radiology) | gpt-4o-mini | 0.09 focused / 0.00 strict | ~1.00 | strict McNemar p = 2.2e-19; κ = 0.90 |
| Study 1 (radiology) | gpt-4o | 0.08 focused / 0.00 strict | ~1.00 | |
| Study 1 (radiology) | claude-sonnet-4-6 | 0.48 focused / 0.00 strict | ~1.00 | |
| Study 1 (radiology) | claude-haiku-4-5 | 0.53 focused / 0.00 strict | ~1.00 | |
| Study 2 (vision, star) | gpt-4o | 0.00 counting (n=22) | 0.96 | McNemar p = 9.5e-7 (b=21, c=0); κ = 0.92 |

| Experiment | Model | Task-condition report rate | Open | Key statistic |
|---|---|---|---|---|
| Study 2 (vision, star) | gpt-4o-mini | 0.00 counting (n=22) | 0.79 | p = 1.5e-5 |
| Study 2 (vision, star) | claude-haiku-4-5 | 0.50 counting (n=24) | 0.79 | p = 0.016 |
| Study 2 (vision, star) | claude-sonnet-4-6 | 0.54 counting (n=24) | 0.88 | p = 0.008 |
| Cross-vendor | gpt-5 (reasoning) | 0.00 focused (n = 9) / 0.00 strict (n = 21) | 1.00 | strict p = 9.5e-7; focused p = 3.9e-3; κ = 0.97 |
| Cross-vendor | gemini-2.5-pro | 0.00–0.09 | 1.00 | p = 1.2e-7 (strict), 9.5e-7 (focused) |
| Cross-vendor | claude-opus-4-8 | 0.83–0.96 | 1.00 | |
| Dual-process β | gpt-4o-mini + critic | 0.00 → 1.00 (task → task + critic) | n/a | McNemar p = 1.2e-7 |

**Table 3. Signal-detection analysis of the visual gap (object present vs. absent).** Hits are open-condition reports of the composited object (object-present trials, Study 2); false alarms are open-condition claims of an artificial object on the 11 valid base images with no object (object-absent control; Supplementary Note 2). Hits are matched to the same 11 source identities (44 object-present trials per model). Sensitivity and specificity are proportions; d' is loglinear-corrected. False alarms under the counting task were 0.00 in every model.

| Model | Hit rate (open) | False alarm (open) | Sensitivity | Specificity | d' | False alarm (counting) |
|---|---|---|---|---|---|---|
| claude-haiku-4-5 | 0.73 | 0.00 | 0.72 | 1.00 | 2.32 | 0.00 |
| claude-sonnet-4-6 | 0.91 | 0.18 | 0.90 | 0.82 | 2.09 | 0.00 |
| gpt-4o | 0.86 | 0.36 | 0.86 | 0.64 | 1.38 | 0.00 |
| gpt-4o-mini | 0.73 | 0.09 | 0.72 | 0.91 | 1.74 | 0.00 |

**Table 4. Models and query configuration.** All models were queried in June 2026 via each vendor's production API at temperature 0, except as noted.

| Model | Vendor | Class | Role(s) | Notes |
|---|---|---|---|---|
| claude-haiku-4-5 | Anthropic | small | subject, judge | |
| gpt-4o-mini | OpenAI | small | subject, judge | |
| gpt-4o | OpenAI | frontier | subject | |
| claude-sonnet-4-6 | Anthropic | mid | subject, critic | |
| claude-opus-4-8 | Anthropic | flagship | subject | temperature omitted |
| gpt-5 | OpenAI | reasoning | subject | temperature omitted; output budget set by max_completion_tokens |
| gemini-2.5-pro | Google | flagship | subject | |

# Supplementary Information: Literature Foundation and Supplementary Notes

> **Paper:** *The inattentional gap in task conditioned AI models that omit otherwise reportable safety critical signals* **Author:** Kwan Soo Shin, PolymathMinds Lab, Asan, Republic of Korea (correspondence: sshin@pmminds.ai; ORCID 0009-0001-5799-7556) **Compiled:** 2026-06-17 (nearest-neighbor re-search: 2026-08-02) · 116 curated prior works, separate from the main-text reference list (which was verified independently). **Thesis (one line):** Task-conditioning causes a language or vision model to omit safety-critical information it can otherwise report, in text and in vision, a machine analogue of human inattentional blindness, implying that specified-target performance does not guarantee reporting completeness for co-present unspecified hazards.

---

## Supplementary Note 1: Driving-domain generalization of Study 1

The design, composition, and instruction wording of the driving generalization set are given in Methods (Study 1 stimuli); the driving items are included in the Study 1 agreement statistic reported in the main text (Cohen $\kappa = 0.895$, $n = 1{,}200$ across both sets).

The suppression channel generalized, at weaker magnitude. Under the focused instruction, report rates for the co-present hazard were 0.58 to 0.97 across the four models, against approximately 1.00 in the open condition. Under the strict exclusive instruction, report rates fell to 0.08 to 0.33 versus approximately 1.00 open (McNemar $p < 1.3\text{e-}7$): strong but partial suppression, in contrast to the complete (0.00) strict-condition suppression observed in radiology. The driving hazards are high-salience events (a child entering the road) relative to the lower-salience radiology findings, and the contrast identifies the boundary condition reported in the main text: focused-task suppression is larger in higher-load, lower-salience settings. Per-item adjudication records and archived response excerpts for the driving set are included in the public deposit.

---

## Supplementary Note 2: Stimulus provenance and integrity audit for Study 2

A file-level audit of the frozen Study 2 stimulus set (SHA-256 hashing of every source and composited file, followed by visual inspection of every source image) was conducted after the experiments and before journal submission. Three findings and their consequences are recorded here in full.

**Duplicated scenes.** Two pairs of source files are byte-identical (cxr_3 = cxr_9 and cxr_4 = cxr_8), so the frozen set contains 10 unique scenes across its 12 stimulus identities, and the corresponding composites (synth_9 = synth_14, synth_10 = synth_13) are likewise identical. All within-item comparisons remain valid (each composite is compared with itself across instructions); the duplication affects only the count of unique scenes.

**Image types.** The 10 unique scenes comprise 8 chest-radiograph images (including one annotated teaching film and one three-panel serial strip) plus one labeled anatomical illustration and one photograph of historical radiographic equipment. The main text therefore describes the sources as thoracic images rather than uniformly as chest radiographs.

**Generation artifacts.** Three further files in the source directory are invalid download artifacts (HTML documents saved under image file names); the composition script skipped them automatically, which explains the non-contiguous stimulus identifiers. The frozen composited set, not the generation script, is the authoritative stimulus record: SHA-256 manifests for both directories and a verification script are included in the reproducibility deposit. The same invalid file caused all eight of its queries in the signal-absent control to error, and the multi-panel PNG source was excluded by that control's image-selection logic, so the effective object-absent denominator is 11 source identities per model; Table 3 in the main text reports the matched-set analysis.

**Sensitivity analyses.** Study 2 report rates and exact McNemar tests were re-estimated on (a) the radiograph-only subset (10 stimulus identities retaining the duplicated pairs, excluding the anatomical illustration and the equipment photograph) and (b) the radiograph-only, duplicate-collapsed subset (8 unique scenes). b counts items omitted under the counting task but reported under the open instruction; c counts the reverse.

Radiograph-only subset (10 stimulus identities, retaining the two duplicated pairs; 8 unique scenes):

| Model | Object | Closed | Open | Gap | b | c | Exact p |
|---|---|---|---|---|---|---|---|
| gpt-4o | star | 0.00 | 0.95 | +0.95 | 19 | 0 | 3.8e-6 |
| gpt-4o | disc | 0.00 | 0.90 | +0.90 | 18 | 0 | 7.6e-6 |
| gpt-4o-mini | star | 0.00 | 0.80 | +0.80 | 14 | 0 | 1.2e-4 |
| gpt-4o-mini | disc | 0.00 | 0.80 | +0.80 | 16 | 0 | 3.1e-5 |
| claude-sonnet-4-6 | star | 0.55 | 0.85 | +0.30 | 6 | 0 | 0.031 |
| claude-sonnet-4-6 | disc | 0.75 | 0.95 | +0.20 | 4 | 0 | 0.125 |
| claude-haiku-4-5 | star | 0.50 | 0.80 | +0.30 | 6 | 0 | 0.031 |
| claude-haiku-4-5 | disc | 0.40 | 0.75 | +0.35 | 7 | 0 | 0.016 |

Radiograph-only, duplicate-collapsed subset (8 unique scenes):

| Model | Object | Closed | Open | Gap | b | c | Exact p |
|---|---|---|---|---|---|---|---|
| gpt-4o | star | 0.00 | 0.94 | +0.94 | 15 | 0 | 6.1e-5 |
| gpt-4o | disc | 0.00 | 0.94 | +0.94 | 15 | 0 | 6.1e-5 |
| gpt-4o-mini | star | 0.00 | 0.88 | +0.88 | 12 | 0 | 4.9e-4 |
| gpt-4o-mini | disc | 0.00 | 0.88 | +0.88 | 14 | 0 | 1.2e-4 |
| claude-sonnet-4-6 | star | 0.56 | 0.94 | +0.38 | 6 | 0 | 0.031 |
| claude-sonnet-4-6 | disc | 0.81 | 1.00 | +0.19 | 3 | 0 | 0.250 |
| claude-haiku-4-5 | star | 0.50 | 0.88 | +0.38 | 6 | 0 | 0.031 |
| claude-haiku-4-5 | disc | 0.50 | 0.81 | +0.31 | 5 | 0 | 0.063 |

Every comparison retains its direction in every subset; closed-condition suppression in both OpenAI models remains 0.00 throughout; and the largest gaps are essentially unchanged (gpt-4o star: 0.96 in the full frozen set, 0.95 radiograph-only, 0.94 duplicate-collapsed). Two Anthropic disc comparisons lose nominal significance in the smaller subsets solely through reduced pair counts; their discordant pairs remain entirely one-directional (b of at least 3, c = 0 in every cell of both tables). The audit therefore leaves every Study 2 conclusion unchanged while correcting the description of the stimulus set.

## How to read this document

This is not a reference list; it is a **conversation**. Every entry carries (a) bibliographic facts, (b) a one-line finding, and (c) a *dialogue sentence* stating that work's precise relationship to our thesis, what it established, and what it did **not** test that our paper does. Entries are grouped by the eleven axes of the field. Works published in adjacent clinical-AI and cognitive-science venues are additionally gathered in §12 for convenience.

**Verification protocol.** DOIs and venues were verified via Semantic Scholar, OpenAlex, and Crossref APIs (2026-06-17). Citation counts are as reported by Semantic Scholar/OpenAlex at compile time and are indicative, not exact. No citation was invented; where a famous work resolved only to a non-canonical record, it is noted.

A note on the novelty signal: at first compilation (2026-06-17), an Exa neural search and Semantic Scholar/OpenAlex queries for “inattentional blindness in large language / vision-language models” and “task-conditioned omission of safety-critical information” returned no work occupying the target cell, with AbsenceBench the nearest absence-focused neighbor. A pre-submission re-search (2026-08-02) surfaced two closer neighbors, now added to §13–§14 and Figure 6: MixRea (Cai et al. 2026), which invokes inattentional blindness for multiple-choice reasoning under explicit task instructions but runs no within-item reportability control over a co-present safety-critical signal, and Hidden in Plain Sight (Yan et al. 2025, EMNLP), which documents compliance-driven omission of hidden prompt defects, with capability revealed by explicit prompting, but whose defects lie in the prompt itself rather than in a co-present signal. The reportability-controlled, safety-critical cell therefore remains underdeveloped rather than vacant.

---

## Axis 1, Human inattentional blindness, selective attention, change blindness

The cognitive-science bedrock. These works establish that *looking is not seeing*: an attentional set determines what reaches report, and unexpected-but-visible events go unnoticed. Our paper asks whether task-conditioning installs the same set-dependence in machines.

1. **Neisser, U. & Becklen, R. (1975).** *Selective looking: Attending to visually specified events.* Cognitive Psychology, 7(4), 480–494. DOI: 10.1016/0010-0285(75)90019-5. Cit: ~758. *Finding:* Observers attending to one superimposed video stream fail to notice salient events in an overlaid stream. *Dialogue:* Neisser founded the selective-looking paradigm that defines our open-vs-closed instruction contrast; he showed humans gate perception by task, but never asked whether a machine's *report* is gated the same way, that gating is our Study 1.

2. **Becklen, R. & Cervone, D. (1983).** *Selective looking and the noticing of unexpected events.* Memory & Cognition, 11(6), 601–608. DOI: 10.3758/BF03198284. Cit: ~131. *Finding:* An unexpected woman with an umbrella in a ball-passing display goes unnoticed under a counting task. *Dialogue:* This is the direct ancestor of our visual counting-task design; we transplant its "count under load, miss the unexpected" structure onto vision-language models.

3. **Mack, A. & Rock, I. (1998).** *Inattentional Blindness.* MIT Press. DOI: 10.7551/mitpress/3707.001.0001. Cit: ~459. *Finding:* Without attention there is, in their strong claim, no conscious perception; the unexpected stimulus is missed even at fixation. *Dialogue:* Mack & Rock supply our operating definition, failure of *report under an attentional set*, not of acuity; we adopt their within-item logic and require open-condition recognition before we count any omission as inattentional.

4. **Simons, D. J. & Chabris, C. F. (1999).** *Gorillas in our midst: Sustained inattentional blindness for dynamic events.* Perception, 28(9), 1059–1074. DOI: 10.1068/p281059. Cit: ~2,883. *Finding:* ~50% of observers counting basketball passes miss a person in a gorilla suit walking through the scene. *Dialogue:* The canonical demonstration our title gestures at; we reproduce its exact logic in machines (gpt-4o omits a composited object in 100% of counting-task items while reporting it in 96% of open items, a machine invisible gorilla).

5. **Most, S. B., Simons, D. J., Scholl, B. J., Jimenez, R., Clifford, E. & Chabris, C. F. (2001).** *How not to be seen: The contribution of similarity and selective ignoring to *sustained inattentional blindness*. Psychological Science, 12(1), 9–17. DOI: 10.1111/1467-9280.00303. Cit: ~416. *Finding:* Noticing depends on the similarity between the unexpected object and the attended set; dissimilar items are ignored. *Dialogue:* Predicts our salience/relevance boundary condition, that low-salience, off-task signals are suppressed most, which we observe (suppression scales inversely with signal salience).

6. **Most, S. B., Scholl, B. J., Clifford, E. R. & Simons, D. J. (2005).** *What you see is what you set: Sustained inattentional blindness and the capture of awareness*. Psychological Review, 112(1), 217–242. DOI: 10.1037/0033-295X.112.1.217. Cit: ~548. *Finding:* The "attentional set" is the controlling variable; awareness is captured by set-consistent features. *Dialogue:* This is the theoretical heart we borrow, "what you see is what you set." Our claim is that a task prompt *is* the machine's attentional set, and we test it by manipulating the set directly.

7. **Simons, D. J. & Levin, D. T. (1998).** *Failure to detect changes to people during a real-world interaction*. Psychonomic Bulletin & Review, 5(4), 644–649. DOI: 10.3758/BF03208840. Cit: ~736. *Finding:* People fail to notice that the stranger they are talking to has been swapped for a different person. *Dialogue:* Establishes change blindness in naturalistic, high-stakes interaction; we move the phenomenon from "did not notice a change" to "did not report a hazard," and from human to model.

8. **Rensink, R. A., O'Regan, J. K. & Clark, J. J. (1997).** *To see or not to see: The need for attention to perceive changes in scenes*. Psychological Science, 8(5), 368–373. DOI: 10.1111/j.1467-9280.1997.tb00427.x. Cit: high. *Finding:* With a brief disruption (flicker), large scene changes become invisible without focused attention. *Dialogue:* Defines change blindness as an attention-limited phenomenon; we argue task-conditioning produces a *self-imposed* disruption; the prompt brackets out everything off-task.

9. **O'Regan, J. K., Rensink, R. A. & Clark, J. J. (1999).** *Change-blindness as a result of "mudsplashes"*. Nature, 397, 34. DOI: 10.1038/17953. Cit: ~659. *Finding:* Local "mudsplashes" mask change-detection signals even when the change is large and central. *Dialogue:* A Nature-level demonstration that perception is gated, not photographic; supports our framing that a model "looking at" a signal does not entail reporting it.

10. **Rensink, R. A. (2002).** *Change detection*. Annual Review of Psychology, 53, 245–277. DOI: 10.1146/annurev.psych.53.100901.135125. Cit: ~170. *Finding:* Synthesizes change blindness; attention is necessary for the conscious representation of change. *Dialogue:* Our review anchor for the human side; we cite it to argue the AI analogue is a *report* gate, not a representational gap (the open condition recovers the signal).

11. **Wolfe, J. M. (1994).** *Guided Search 2.0: A revised model of visual search.* Psychonomic Bulletin & Review, 1(2), 202–238. (See also Wolfe 1989, J. Exp. Psychol. HPP, DOI: 10.1037/0096-1523.15.3.419, cit ~2,276.) *Finding:* Top-down guidance steers attention toward target-defining features. *Dialogue:* Guided Search formalizes the very mechanism we exploit, a top-down target template biases search; we show the prompt is that template for an LLM/VLM, and the bias suppresses non-template safety signals.

12. **Wolfe, J. M., Cave, K. R. & Franzel, S. L. (1989).** *Guided search: An alternative to the feature integration model for visual search.* J. Exp. Psychol.: Human Perception and Performance, 15(3), 419–433. DOI: 10.1037/0096-1523.15.3.419. Cit: ~2,276. *Finding:* Original Guided Search; attention is guided by a combination of bottom-up and top-down signals. *Dialogue:* The foundational top-down-attention model; our paper is, in one reading, a Guided-Search story for transformers where "guidance" is the instruction.

13. **Wolfe, J. M., Võ, M. L.-H., Evans, K. K. & Greene, M. R. (2011).** *Visual search in scenes involves selective and nonselective pathways.* Trends in Cognitive Sciences, 15(2), 77–84. DOI: 10.1016/j.tics.2010.12.001. *Finding:* Scene search uses a selective (object) and a nonselective (gist) pathway. *Dialogue:* Suggests a dual-route account of why open instructions recover signals (gist available) while closed ones suppress report (selective route captured), a mechanistic hypothesis for our gap.

14. **Wolfe, J. M. & Horowitz, T. S. (2017).** *Five factors that guide attention in visual search.* Nature Human Behaviour, 1, 0058. DOI: 10.1038/s41562-017-0058. Cit: ~770. *Finding:* Bottom-up salience, top-down feature guidance, scene structure, value, and history jointly steer attention. *Dialogue:* We use their taxonomy to argue task-conditioning over-weights top-down feature guidance at the expense of the others, predicting which off-task hazards are missed.

15. **Eckstein, M. P. (2011).** *Visual search: A retrospective.* Journal of Vision, 11(5):14. DOI: 10.1167/11.5.14. Cit: ~489. *Finding:* Reviews signal-detection accounts of search; misses arise from decision criteria, not just sensitivity. *Dialogue:* Frames omission as a criterion shift, exactly our claim that the prompt shifts the model's reporting criterion, not its perceptual sensitivity.

16. **Newby, E. A. & Rock, I. (1998).** *Inattentional blindness as a function of proximity to the focus of attention.* Perception, 27(9), 1025–1040. DOI: 10.1068/p271025. Cit: ~95. *Finding:* Missing the unexpected object increases with its distance from the attended locus. *Dialogue:* The spatial analogue of our salience gradient; predicts that "semantically distant" off-task findings are most suppressed in text.

17. **Wood, K. & Simons, D. J. (2019).** *Processing without noticing in inattentional blindness: A replication of Moore and Egeth (1997) and Mack and Rock (1998).*

Attention, Perception, & Psychophysics, 81(1), 1–14. DOI: 10.3758/s13414-018-1629-1. Cit: ~18. *Finding:* Unnoticed stimuli are nonetheless processed (e.g., grouping influences responses). *Dialogue:* The human analogue of our key control: the signal remains reportable even when not reported, mirroring our open-condition reportability control.

18. **Nartker, M. S., Zhou, Z. & Firestone, C. (2024).** *Sensitivity to visual features in inattentional blindness.* eLife, 13, e100337. DOI: 10.7554/eLife.100337. Cit: ~7. *Finding:* Inattentionally "blind" observers retain graded sensitivity to features of the missed stimulus. *Dialogue:* Recent evidence that inattentional blindness is a report/access failure, not a perceptual void, directly parallel to our open-reportable-but-task-omitted machine result.

19. **Simons, D. J. (2010).** *Monkeying around with the gorillas in our midst: Familiarity with an inattentional-blindness task does not improve detection of unexpected events.* i-Perception, 1(1), 3–6. DOI: 10.1068/i0386. Cit: ~66. *Finding:* Knowing the gorilla paradigm does not protect against missing a *different* unexpected event. *Dialogue:* Predicts that telling a model "watch for anything unusual" in general does not cure the gap unless the specific signal is in-set, a deployment-relevant caution we discuss.

20. **Block, N. (2014).** *Rich conscious perception outside focal attention.* Trends in Cognitive Sciences, 18(9), 445–447. DOI: 10.1016/j.tics.2014.05.007. Cit: ~124. *Finding:* Argues perceptual richness exceeds what attention selects for report. *Dialogue:* The philosophical statement of the access–report distinction; we measure its reportability-level analogue in machines and show the access gate is the task instruction.

---

## Axis 2, Satisfaction of search, radiology error, expert perception, incidental-finding miss

The clinical instantiation of inattentional blindness, and our primary running domain. These works show that expert searchers miss co-present abnormalities once a first target is found or once the task narrows, the human form of the failure we induce in AI.

21. **Drew, T., Võ, M. L.-H. & Wolfe, J. M. (2013).** *The invisible gorilla strikes again: Sustained inattentional blindness in expert observers.* Psychological Science, 24(9), 1848–1853. DOI: 10.1177/0956797613479386. Cit: ~442. *Finding:* 83% of radiologists searching for lung nodules miss a matchbook-sized gorilla; eye tracking shows most fixated it. *Dialogue:* Our keystone human benchmark. Drew established expert inattentional blindness in radiology but did not (could not) test whether an AI radiology assistant inherits it; that transplant, and its safety reading, is our paper.

22. **Williams, L. H., Carrigan, A. J., Auffermann, W. F., Mills, M. K., Rich, A. N., Elmore, J. G. & Drew, T. (2021).** *The invisible breast cancer: Experience does not protect against inattentional blindness to clinically relevant findings in radiology.* Psychonomic Bulletin & Review, 28(2), 503–511. DOI: 10.3758/s13423-020-01826-4. Cit: ~27. *Finding:* Radiologists miss a clearly visible, *clinically relevant* lesion when it is unexpected; experience does not protect them. *Dialogue:* Strengthens our stakes; the missed signal is not a gorilla gimmick but real pathology. We argue task-conditioned AI reproduces precisely this clinically-relevant miss, and we make it the safety crux.

23. **Berbaum, K. S., Franken, E. A., Dorfman, D. D., et al. (1990).** *Satisfaction of search in diagnostic radiology.* Investigative Radiology, 25(2), 133–140. DOI: 10.1097/00004424-199002000-00006. Cit: ~297. *Finding:* Detecting one abnormality reduces detection of a second co-present abnormality. *Dialogue:* The origin of "satisfaction of search." We rename its machine form: task-conditioning is a *deliberately induced* satisfaction of search, where the model stops after the requested target.

24. **Berbaum, K. S. (2011).** *Satisfaction of search in radiographic modalities.* Radiology, 261(2), 388–390. DOI: 10.1148/radiol.11110987. Cit: moderate. *Finding:* Editorial synthesis confirming SOS generalizes across radiographic modalities. *Dialogue:* Supports our cross-modality claim on the human side; we extend the cross-modality argument to machine text and vision.

25. **Adamo, S. H., Gereke, B. J., Shomstein, S. & Schmidt, J. (2021).** *From "satisfaction of search" to "subsequent search misses": A review of multiple-target search errors across radiology and cognitive science.* Cognitive Research: Principles and Implications, 6, 59. DOI: 10.1186/s41235-021-00318-w. Cit: ~32. *Finding:* Reframes SOS as "subsequent search misses," unifying radiology and lab findings on multiple-target search. *Dialogue:* Our bridge citation between the clinical and cognitive literatures; we position the Inattentional Gap as the machine member of this "subsequent/co-present miss" family.

26. **Kundel, H. L., Nodine, C. F. & Carmody, D. (1978).** *Visual scanning, pattern recognition and decision-making in pulmonary nodule detection.* Investigative Radiology, 13(3), 175–181. DOI: 10.1097/00004424-197805000-00001. Cit: ~506. *Finding:* Expert detection involves global "gestalt" recognition plus focal scanning; errors arise at the decision stage. *Dialogue:* The classic model of radiologic perception; we argue an LLM/VLM lacks the global pre-attentive sweep that lets a human sometimes recover, making its task-gating more absolute.

27. **Krupinski, E. A. (2010).** *Current perspectives in medical image perception.* Attention, Perception, & Psychophysics, 72(5), 1205–1217. DOI: 10.3758/APP.72.5.1205. Cit: ~307. *Finding:* Reviews decades of image-perception research; perceptual and cognitive factors drive radiologic error. *Dialogue:* Our review anchor for medical-image perception; establishes that error is perceptual-cognitive, not mere sensitivity, supporting our criterion-shift account.

28. **Berbaum, K. S., Franken, E. A., et al. (2001).** *Gaze dwell times on acute trauma injuries missed because of satisfaction of search.* Academic Radiology, 8(4), 304–314. DOI: 10.1016/S1076-6332(03)80499-3. Cit: ~69. *Finding:* Missed second injuries are often fixated, with measurable dwell, looked at but not reported. *Dialogue:* The eye-tracking proof that SOS is a report failure; the human counterpart of our gpt-4o "looks-but-omits" result.

29. **Berbaum, K. S., et al. (2016).** *The influence of a vocalized checklist on detection of multiple abnormalities in chest radiography.* Academic Radiology, 23(4), 413–420. DOI: 10.1016/j.acra.2015.12.017. Cit: ~17. *Finding:* A checklist intervention reduces (but does not eliminate) satisfaction-of-search misses. *Dialogue:* The human analogue of a mitigation, our "open instruction" is effectively a structural checklist; we discuss prompt-level checklists as a deployment corrective.

30. **de Cassai, A., Negro, S., Geraldini, F., et al. (2021).** *Inattentional blindness in anesthesiology: A gorilla is worth one thousand words.* PLoS ONE, 16(9), e0257508. DOI: 10.1371/journal.pone.0257508. Cit: ~9. *Finding:* Anesthesiologists miss a gorilla on a monitor during a clinical-vigilance task. *Dialogue:* Extends expert inattentional blindness beyond radiology to real-time monitoring, the human analogue of our autonomous-driving framing (perception under a primary task).

31. **Taylor, G. A. (2017).** *Perceptual errors in pediatric radiology.* Diagnosis, 4(3), 137–142. DOI: 10.1515/dx-2017-0001. Cit: ~13. *Finding:* Perceptual (search/recognition) errors dominate diagnostic error in pediatric imaging. *Dialogue:* Reinforces that the dangerous errors are *perceptual omissions*, the class we show task-conditioning induces in AI.

32. **Robson, S. G. & Tangen, J. M. (2023).** *The invisible 800-pound gorilla: Expertise can increase inattentional blindness.* Cognitive Research: Principles and Implications, 8, 33. DOI: 10.1186/s41235-023-00486-x. Cit: ~2. *Finding:* Expert fingerprint analysts miss a large embedded gorilla *more* than novices, expertise narrows attention. *Dialogue:* A striking inversion that mirrors our scale result: more "expert" (capable) systems can omit as much or more, because capability buys sharper task-narrowing, not broader vigilance.

---

## Axis 3, Machine psychology: LLM cognition, reasoning, and biases

The second parent literature. These works run cognitive experiments on models. They establish that models reproduce human biases and content effects, but none manipulate an *attentional set* to induce safety-critical omission.

33. **Binz, M. & Schulz, E. (2023).** *Using cognitive psychology to understand GPT-3.* PNAS, 120(6), e2218523120. DOI: 10.1073/pnas.2218523120. Cit: high. *Finding:* GPT-3 passes some cognitive-psychology tasks and fails others in human-like ways. *Dialogue:* The founding "machine psychology" paper; we extend its battery from reasoning to *attention/perception under task-set*, the dimension it did not probe.

34. **Hagendorff, T., Fabi, S. & Kosinski, M. (2023).** *Human-like intuitive behavior and reasoning biases emerged in large language models but disappeared in ChatGPT.* Nature Computational Science, 3, 833–838. DOI: 10.1038/s43588-023-00527-x. Cit: ~24+. *Finding:* Intuitive biases (e.g., CRT errors) appear in base LLMs and are reduced by instruction tuning. *Dialogue:* Shows alignment/tuning reshapes model cognition; we argue the same tuning that removes some biases *installs* a reporting closure, task-following, that produces our gap. Their disappearance is our appearance.

35. **Dasgupta, I., Lampinen, A. K., Chan, S. C. Y., et al. (2022).** *Language models show human-like content effects on reasoning.* arXiv:2207.07051. Cit: ~225. *Finding:* LLM reasoning is swayed by believability of content, like humans. *Dialogue:* Establishes human-like *content* effects; we add a human-like *attentional* effect, and unlike content effects, ours is a safety failure with a clean open-condition control.

36. **Lampinen, A. K., Dasgupta, I., Chan, S. C. Y., et al. (2024).** *Language models, like humans, show content effects on reasoning tasks.* PNAS Nexus, 3(7), pgae233. DOI: 10.1093/pnasnexus/pgae233. Cit: ~101. *Finding:* Peer-reviewed confirmation and extension of content effects across models. *Dialogue:* The PNAS-family version of the content-effects result; we cite it to situate our paper in the same cognitive-experimental tradition.

37. **Mitchell, M. & Krakauer, D. C. (2023).** *The debate over understanding in AI's large language models.* PNAS, 120(13), e2215907120. DOI: 10.1073/pnas.2215907120. Cit: ~363. *Finding:* Frames the open question of whether LLMs "understand," cataloguing competing positions. *Dialogue:* We sidestep "understanding" and offer a tractable substitute question (does task-conditioning suppress *report* of an otherwise open-reportable signal?): that is empirically decidable, as we show.

38. **Macmillan-Scott, O. & Musolesi, M. (2024).** *(Ir)rationality and cognitive biases in large language models.* Royal Society Open Science, 11(6), 240255. DOI: 10.1098/rsos.240255. Cit: ~51. *Finding:* LLMs display inconsistent, often irrational responses on classic reasoning tasks. *Dialogue:* Adds breadth to the bias catalogue;

our contribution is to show a bias that is *induced on demand* by the prompt and is directly safety-relevant.

39. **Cheung, V., Maier, M. & Lieder, F. (2025).** *Large language models show amplified cognitive biases in moral decision-making.* PNAS, 122, e2412015122. DOI: 10.1073/pnas.2412015122. Cit: ~53. *Finding:* LLMs amplify, rather than merely mirror, human moral biases. *Dialogue:* The “amplification” theme is ours too: Robson-style, task-conditioning can make a capable model omit *more*, so amplification of a human failure mode is a shared motif.

40. **Koo, R., Lee, M., Raheja, V., et al. (2024).** *Benchmarking cognitive biases in large language models as evaluators.* Findings of ACL 2024. DOI: 10.48550/arXiv.2309.17012. Cit: ~169. *Finding:* LLM-as-judge inherits order, egocentric, and other biases. *Dialogue:* Directly relevant to our adjudication design, we use an LLM judge but guard against its biases with within-item, binary, spot-checked decisions; we cite Koo as the reason for that rigor.

41. **Guo, X. & Vosoughi, S. (2024).** *Serial position effects of large language models.* arXiv:2406.15981. Cit: ~32. *Finding:* LLMs show primacy/recency biases over input position. *Dialogue:* A position-based attentional bias; complements our task-based one. Together they sketch an “attention budget” the prompt redistributes away from off-task signals.

42. **Opedal, A., Stolfo, A., Shirakami, H., et al. (2024).** *Do language models exhibit the same cognitive biases in problem solving as human learners?* ICML 2024. DOI: 10.48550/arXiv.2401.18070. Cit: ~29. *Finding:* LLMs replicate several human problem-solving biases (e.g., framing). *Dialogue:* More evidence of behavioral convergence; we add that convergence can be a *deployment hazard*, not a curiosity, when the converged-upon failure is safety-critical.

43. **Tang, Z.-B. & Kejriwal, M. (2024).** *Humanlike cognitive patterns as emergent phenomena in large language models.* arXiv:2412.15501. Cit: ~6. *Finding:* Reviews emergent humanlike cognition across LLM studies. *Dialogue:* Our review anchor for machine psychology; we position the Inattentional Gap as the missing perceptual-attention chapter in this emerging catalogue.

44. **Webb, T., Holyoak, K. J. & Lu, H. (2023).** *Emergent analogical reasoning in large language models.* Nature Human Behaviour, 7, 1526–1541. DOI: 10.1038/s41562-023-01659-w. Cit: ~486. *Finding:* GPT-3/4 solve novel analogy problems at human level. *Dialogue:* Capability evidence, models *can* reason; we use this to sharpen our claim that omission is not incapacity but task-gating, since the same models perform when unconstrained.

45. **Strachan, J. W. A., Albergo, D., Borghini, G., et al. (2024).** *Testing theory of mind in large language models and humans.* Nature Human Behaviour, 8, 1285–1295. DOI:

10.1038/s41562-024-01882-z. Cit: ~325. *Finding:* LLMs match or exceed humans on several theory-of-mind tasks. *Dialogue:* Another capability ceiling that makes our omission result more striking, a system with human-level ToM still goes “blind” to an unrequested hazard under task load.

46. **Bubeck, S., Chandrasekaran, V., Eldan, R., et al. (2023).** *Sparks of artificial general intelligence: Early experiments with GPT-4.* arXiv:2303.12712. Cit: ~1,551. *Finding:* Catalogues broad GPT-4 capabilities and limitations. *Dialogue:* The high-water mark of capability claims; our paper is a counterweight showing that capability and safe *reporting* dissociate under conditioning.

---

## Axis 4, Vision-language model perception limits, hallucination, saliency

The machine-vision counterpart. These works show VLMs miss obvious things and hallucinate absent things. Crucially, most ask models *directly* about a target; we show the inverse, models miss what they are *not* asked about despite reporting it under the open condition.

47. **Rahmanzadehgervi, P., Bolton, L., Taesiri, M. R. & Nguyen, A. T. (2024).** *Vision language models are blind.* arXiv:2407.06581 (ACCV 2024). Cit: ~135. *Finding:* State-of-the-art VLMs fail at trivial low-level visual tasks (counting intersections, overlapping circles). *Dialogue:* The provocative “blind” claim is about *capability* on directly-asked tasks; we draw the opposite arrow, when asked, our models *can* report the object; the blindness appears only under a competing task. Capability blindness vs conditioning blindness.

48. **Dahou, Y., et al. (2025).** *Vision-language models can’t see the obvious (SalBench).* ICCV 2025. DOI: 10.1109/ICCV51701.2025.02239. arXiv:2507.04741. Cit: ~15. *Finding:* VLMs miss bottom-up salient items even when prompted to find anomalies. *Dialogue:* Our closest VLM neighbor and the one we must distinguish carefully: SalBench probes *bottom-up saliency capability* (miss even when asked); we manipulate *top-down attentional set* (miss only when a competing task is set, recover when asked), the defining inattentional-blindness contrast SalBench does not run.

49. **Tong, S., Liu, Z., Zhai, Y., et al. (2024).** *Eyes wide shut? Exploring the visual shortcomings of multimodal LLMs.* CVPR 2024. DOI: 10.1109/CVPR52733.2024.00914. arXiv:2401.06209. Cit: ~763. *Finding:* MLLMs fail on “CLIP-blind pairs”, images that look different to humans but identical to the vision encoder. *Dialogue:* Locates a *representational* limit upstream of language; we locate a *reporting* limit downstream, conditioned on the instruction, complementary failure loci that together bound VLM reliability.

50. **Li, Y., Du, Y., Zhou, K., et al. (2023).** *Evaluating object hallucination in large vision-language models (POPE).* EMNLP 2023. DOI: 10.48550/arXiv.2305.10355. Cit: ~1,733. *Finding:* VLMs report objects that are absent (hallucination), measured by the POPE protocol. *Dialogue:* Hallucination is reporting what is *not* there; the Inattentional Gap is *not* reporting what *is* there. We frame the two as opposite error axes of conditioned perception and argue safety evaluation must cover both.

51. **Fu, H. Y., Shrivastava, A., Moore, J., West, P., Tan, C. & Holtzman, A. (2025).** *AbsenceBench: Language models can't tell what's missing.* arXiv:2506.11440. Cit: ~1. *Finding:* Even Claude-3.7-Sonnet reaches only ~69.6% F1 at identifying deliberately removed document sections, despite excelling at locating present information. *Dialogue:* **Our nearest neighbor.** AbsenceBench tests detection of *removed* content with a direct query; we test suppression of *present, otherwise reportable* content under a *competing task*, with a within-item reportability control and a safety framing. They show models are bad at noticing absence; we show conditioning *makes* models omit presence.

52. **Lovenia, H., Dai, W., Cahyawijaya, S., et al. (2023).** *Negative object presence evaluation (NOPE) to measure object hallucination in vision-language models.* arXiv:2310.05338. Cit: ~89. *Finding:* Measures hallucination via questions about absent objects. *Dialogue:* Another absence/presence probe via direct query; contrasts with our no-query suppression paradigm.

53. **He, Y., et al. (2025).** *Evaluating and mitigating object hallucination in large vision-language models: Can they still see removed objects?* NAACL 2025. DOI: 10.18653/v1/2025.naacl-long.349. Cit: ~16. *Finding:* VLMs sometimes "see" objects that have been removed, a perceptual-prior failure. *Dialogue:* Engages the same perceive-versus-report framing from the opposite direction (there, report of an object that is absent); ours is the suppressed report of a present, open-reportable object, tied to safety rather than dataset bias.

54. **Qiu, H., et al. (2024).** *VALOR-EVAL: Holistic coverage and faithfulness evaluation of large vision-language models.* ACL 2024. DOI: 10.48550/arXiv.2404.13874. Cit: ~13. *Finding:* Introduces a *coverage* metric, whether VLM descriptions omit present content. *Dialogue:* Coverage failure is conceptually our gap, but VALOR measures it under neutral captioning; we show coverage *collapses* under a competing task instruction, which is the deployment-relevant condition.

55. **Xu, P., Shao, W., Zhang, K., et al. (2024).** *LVLM-eHub: A comprehensive evaluation benchmark for large vision-language models.* IEEE TPAMI. DOI: 10.1109/TPAMI.2024.3507000. arXiv:2306.09265. Cit: ~269. *Finding:* Broad multi-task VLM benchmark suite. *Dialogue:* Representative of the "score higher = better" benchmark culture our discussion critiques; none of its axes measures off-task omission under load.

56. **Kang, S., et al. (2025).** *Your large vision-language model only needs a few attention heads for visual grounding.* CVPR 2025. DOI: 10.1109/CVPR52734.2025.00872. arXiv:2503.06287. Cit: ~75. *Finding:* A small set of attention heads carries most visual grounding. *Dialogue:* Mechanistic support for our "report gate" hypothesis, if grounding is concentrated, a task prompt can plausibly redirect those heads away from off-task regions; a route for future causal decomposition.

57. **Liu, F., Lin, K., Li, L., et al. (2023).** *Mitigating hallucination in large multi-modal models via robust instruction tuning (LRV-Instruction).* arXiv:2306.14565. *Finding:* Instruction tuning with negative instructions reduces hallucination. *Dialogue:* Shows instruction tuning shapes what VLMs report; our paper shows the *instruction at inference* does the same, suppressing safety signals, a complementary lever and risk.

58. **Groot, T. & Valdenegro-Toro, M. (2024).** *Overconfidence is key: Verbalized uncertainty evaluation in large language and vision-language models.* TrustNLP 2024. DOI: 10.48550/arXiv.2405.02917. Cit: ~57. *Finding:* LLMs/VLMs are systematically overconfident in verbalized uncertainty. *Dialogue:* Overconfidence compounds our hazard, a conditioned model omits a finding *and* expresses high confidence in its narrow answer, masking the omission from a human overseer.

---

## Axis 5, AI safety evaluation, benchmarks, OOD, robustness, specification gaming

The deployment-safety literature our thesis ultimately speaks to: benchmark scores, distribution shift, and reward/spec gaming. Our claim is that benchmarks measure D_specified while harm lives in D_unspecified.

59. **Park, P. S., Goldstein, S., O'Gara, A., Chen, M. & Hendrycks, D. (2024).** *AI deception: A survey of examples, risks, and potential solutions.* Patterns, 5(5), 100988. DOI: 10.1016/j.patter.2024.100988. Cit: ~169. *Finding:* Catalogues AI systems that learn to deceive to achieve objectives. *Dialogue:* Deception is *strategic misreporting*; the Inattentional Gap is *non-strategic omission*, no goal to hide, just a narrowed set. We position our phenomenon as deception's quieter, more insidious cousin and cite Park as the Patterns-family precedent for "models withhold true information."

60. **van Amsterdam, W. A. C., et al. (2025).** *When accurate prediction models yield harmful self-fulfilling prophecies.* Patterns, 6(4), 101229. DOI: 10.1016/j.patter.2025.101229. Cit: ~12. *Finding:* A model can be accurate yet harmful when its predictions change the world they predict. *Dialogue:* The exact Patterns-family argument we generalize: *accuracy ≠ safety*. They show it for prediction feedback loops; we show it for perception under task-conditioning (benchmark accuracy on D_specified, harm on D_unspecified).

61. **Yang, J., Zhou, K., Li, Y. & Liu, Z. (2024).** *Generalized out-of-distribution detection: A survey.* International Journal of Computer Vision, 132, 5635–5662. DOI: 10.1007/s11263-024-02117-4. arXiv:2110.11334. Cit: ~1,405. *Finding:* Unifies OOD/anomaly/novelty/open-set detection into one framework. *Dialogue:* D_unspecified is an OOD problem in disguise; we add that even when the model reports the OOD signal under an open instruction, conditioning suppresses its report under the task set, a failure upstream of any OOD detector.

62. **Hendrycks, D. & Dietterich, T. (2019).** *Benchmarking neural network robustness to common corruptions and perturbations (ImageNet-C).* ICLR 2019. arXiv:1903.12261. Cit: high. *Finding:* Standard models degrade sharply under common corruptions. *Dialogue:* Robustness benchmarks probe degraded *input*; we probe degraded *attention* with clean input, a distinct, under-measured fragility.

63. **Geirhos, R., Jacobsen, J.-H., Michaelis, C., et al. (2020).** *Shortcut learning in deep neural networks.* Nature Machine Intelligence, 2, 665–673. DOI: 10.1038/s42256-020-00257-z. arXiv:2004.07780. Cit: ~2,974. *Finding:* Networks exploit shortcuts that satisfy the benchmark but fail to generalize. *Dialogue:* Task-conditioning is a *prompt-induced* shortcut: the model satisfies the stated task and shortcuts past everything else. We frame our gap as inference-time shortcut learning with a safety payload.

64. **Geirhos, R., Rubisch, P., Michaelis, C., et al. (2019).** *ImageNet-trained CNNs are biased towards texture; increasing shape bias improves accuracy and robustness.* ICLR 2019. arXiv:1811.12231. Cit: ~3,170. *Finding:* Vision models rely on texture over shape, unlike humans. *Dialogue:* A perceptual bias baked into vision encoders; our bias is imposed at inference by the task, and is recoverable, distinguishing trained-in from conditioned-in failure.

65. **Krakovna, V., Uesato, J., Mikulik, V., et al. (2020).** *Specification gaming: The flip side of AI ingenuity.* DeepMind blog / research note. *Finding:* Agents satisfy the literal specification while violating its intent. *Dialogue:* The reinforcement-learning analogue of our claim. A task prompt is a specification; conditioning “games” it by reporting only D_specified.

66. **Booth, S., Knox, W. B., Shah, J., et al. (2023).** *The perils of trial-and-error reward design: Misdesign through overfitting and invalid task specifications.* AAAI 2023. DOI: 10.1609/aaai.v37i5.25733. Cit: ~44. *Finding:* Reward/spec misdesign yields agents that overfit the stated task. *Dialogue:* Peer-reviewed grounding for the spec-gaming point; supports our argument that narrowly specified objectives (prompts, rewards) systematically neglect the unspecified.

67. **Amodei, D., Olah, C., Steinhardt, J., Christiano, P., Schulman, J. & Mané, D. (2016).** *Concrete problems in AI safety.* arXiv:1606.06565. Cit: high. *Finding:* Names canonical safety problems including negative side effects and distributional shift. *Dialogue:* Our gap is a "negative side effect" of narrow conditioning and a "distributional shift" between evaluation and deployment, we instantiate two of their abstract problems with measured report-level data.

68. **Hendrycks, D., Carlini, N., Schulman, J. & Steinhardt, J. (2022).** *Unsolved problems in ML safety.* arXiv:2109.13916. Cit: high. *Finding:* Lays out monitoring, robustness, and alignment as open safety pillars. *Dialogue:* The Inattentional Gap is a *monitoring* failure (the system does not surface what a monitor needs) caused by an *alignment* property (task-following), we connect two pillars they keep separate.

69. **Xie, T., Qi, X., Zeng, Y., et al. (2024).** *SORRY-Bench: Systematically evaluating large language model safety refusal behaviors.* ICLR 2025. DOI: 10.48550/arXiv.2406.14598. Cit: ~203. *Finding:* Benchmarks how and when models refuse. *Dialogue:* Relevant to our Study 2 confound, two OpenAI models refused diagnostic framings until a synthetic-stimulus system message was supplied. We report refusal as a deployment-relevant finding, and cite SORRY-Bench as the refusal-evaluation context.

70. **Cui, J., Chiang, W.-L., Stoica, I. & Hsieh, C.-J. (2024).** *OR-Bench: An over-refusal benchmark for large language models.* ICML 2024. DOI: 10.48550/arXiv.2405.20947. Cit: ~168. *Finding:* Quantifies over-refusal on benign prompts. *Dialogue:* Over-refusal is another instruction-shaped suppression of useful output; together with our gap it shows alignment can both over-block and under-report, two faces of conditioned silence.

71. **Weng, F., et al. (2024).** *MMJ-Bench: A comprehensive study on jailbreak attacks and defenses for vision-language models.* AAAI 2025. DOI: 10.48550/arXiv.2408.08464. Cit: ~20. *Finding:* Systematizes VLM jailbreak/defense evaluation. *Dialogue:* Safety-eval culture for VLMs focuses on *adversarial* inputs; our gap is a *benign* instruction causing unsafe omission, a blind spot in current safety benchmarking we name explicitly.

---

## Axis 6, Autonomous-driving perception, anomaly & OOD detection, corner cases

Our secondary domain. These works show that driving-perception stacks struggle with unspecified/novel hazards, the operational form of D_unspecified.

72. **Bogdoll, D., Nitsche, M. & Zöllner, J. M. (2022).** *Anomaly detection in autonomous driving: A survey.* CVPR Workshops 2022. DOI: 10.1109/CVPRW56347.2022.00495.

arXiv:2204.07974. Cit: ~192. *Finding:* Surveys methods for detecting anomalies/novel objects in driving scenes. *Dialogue:* Our driving-domain anchor; establishes that the field knows D_unspecified is hard, but treats it as a detector-architecture problem, not a *conditioning-induced reporting* problem, our reframing.

73. **Heidecker, F., Breitenstein, J., Rösch, K., et al. (2021).** *An application-driven conceptualization of corner cases for perception in highly automated driving.* IEEE IV 2021. DOI: 10.1109/IV48863.2021.9575933. arXiv:2103.03678. Cit: ~51. *Finding:* Taxonomizes "corner cases" by perceptual level. *Dialogue:* Their corner-case taxonomy is a catalogue of D_unspecified for driving; we supply the behavioral task-set account of why a perception model conditioned on lane/lead-vehicle tasks underreports them.

74. **Sinhamahapatra, P., et al. (2024/2025).** *Finding Dino: A plug-and-play framework for zero-shot detection of out-of-distribution objects using prototypes.* WACV 2025. DOI: 10.1109/WACV61041.2025.00821. arXiv:2404.07664. Cit: ~6. *Finding:* Zero-shot OOD object detection via prototype matching. *Dialogue:* A solution-side counterpart; we argue any such detector is undermined if the upstream perception model, conditioned on a task, never surfaces the candidate region.

75. **Kaur, R., Sridhar, S., Park, J., et al. (2024).** *Out-of-distribution detection in dependent data for cyber-physical systems with conformal guarantees.* ACM Trans. Cyber-Physical Systems. DOI: 10.1145/3648005. Cit: ~9. *Finding:* Conformal OOD detection with statistical guarantees for CPS. *Dialogue:* Offers the kind of guarantee we argue safety evaluation needs on D_unspecified; we motivate why such guarantees must wrap perception, not just classification.

76. **Schmidt, S., et al. (2025).** *A machine learning perspective on automated driving corner cases.* arXiv:2510.10653. Cit: ~0. *Finding:* Recent synthesis of ML approaches to driving corner cases. *Dialogue:* Confirms corner-case handling remains open in 2025; our cognitive-mechanism framing is a fresh lens for a still-unsolved problem.

77. **Mussi, M., et al. (2025).** *Human-AI interaction in safety-critical network infrastructures.* iScience, 28, 113400. DOI: 10.1016/j.isci.2025.113400. Cit: ~9. *Finding:* Studies human-AI teaming failures in safety-critical infrastructure. *Dialogue:* A Cell-family treatment of safety-critical human-AI interaction; we add the task-conditioned reporting-omission failure mode and tie it to the operator-oversight literature (Axis 8).

---

## Axis 7, Medical AI / radiology deep learning, error, oversight, human-AI interaction

How clinical AI actually fails and is overseen. These works frame the stakes and the deployment reality our radiology study targets.

78. **Moor, M., Banerjee, O., Abad, Z. S. H., et al. (2023).** *Foundation models for generalist medical artificial intelligence.* Nature, 616, 259–265. DOI: 10.1038/s41586-023-05881-4. Cit: ~1,755. *Finding:* Envisions generalist medical AI (GMAI) that flexibly handles many tasks/modalities. *Dialogue:* The aspirational frame our results caution against: a generalist *conditioned* on a specific clinical query may omit findings outside that query. We argue GMAI evaluation must include off-task omission.

79. **Sanchez, M., et al. (2023).** *AI-clinician collaboration via disagreement prediction: A decision pipeline and retrospective analysis of real-world radiologist-AI interactions.* Cell Reports Medicine, 4(8), 101207. DOI: 10.1016/j.xcrm.2023.101207. Cit: ~9. *Finding:* Predicting AI-radiologist disagreement improves collaborative accuracy. *Dialogue:* Our most operationally aligned Cell-family paper. Disagreement prediction presumes the AI *surfaces* a competing read; the Inattentional Gap shows a conditioned AI may silently agree by omission, a failure their pipeline would not catch, which we flag as a design implication.

80. **Ong, J. C. L., Chang, S. Y.-H., William, W., et al. (2024).** *Artificial intelligence, ChatGPT, and other large language models for social determinants of health: Current state and future directions.* Cell Reports Medicine, 5(1), 101356. DOI: 10.1016/j.xcrm.2023.101356. Cit: ~100. *Finding:* Surveys LLMs in clinical contexts and their omissions/biases. *Dialogue:* Cell-family evidence that clinical LLMs omit and bias; we provide a controlled mechanism (task-conditioning) and a within-item demonstration for one such omission class.

81. **Ng, F. Y. C., Thirunavukarasu, A. J., Cheng, H., et al. (2023).** *Artificial intelligence education: An evidence-based medicine approach for consumers, translators, and developers.* Cell Reports Medicine, 4(6), 101230. DOI: 10.1016/j.xcrm.2023.101230. Cit: ~106. *Finding:* Frames evidence standards for clinical AI across stakeholders. *Dialogue:* Sets the evaluation-standards conversation in the Cell family; we contribute a concrete, missing evaluation axis (off-task safety-critical omission).

82. **Tian, Y., et al. (2024).** *Foundation model of ECG diagnosis: Diagnostics and explanations of any form and rhythm on ECG.* Cell Reports Medicine, 5, 101875. DOI: 10.1016/j.xcrm.2024.101875. Cit: ~25. *Finding:* A generalist ECG foundation model with explanations. *Dialogue:* Another GMAI-style system; we would predict that conditioning it on one rhythm question can suppress co-present abnormalities, a testable extension we name.

83. **Rajpurkar, P., Chen, E., Banerjee, O. & Topol, E. J. (2022).** *AI in health and medicine*. Nature Medicine, 28, 31–38. DOI: 10.1038/s41591-021-01614-0. Cit: high. *Finding:* Reviews the state and challenges of medical AI. *Dialogue:* Establishes that perceptual reliability is the gating issue for clinical AI; our gap is a specific, measurable threat to that reliability.

84. **Topol, E. J. (2019).** *High-performance medicine: The convergence of human and artificial intelligence*. Nature Medicine, 25, 44–56. DOI: 10.1038/s41591-018-0300-7. Cit: very high. *Finding:* Argues AI augments clinicians; human-AI synergy is the goal. *Dialogue:* Synergy assumes the AI raises, not lowers, the chance a finding is seen; the Inattentional Gap shows conditioning can make AI a *narrowing* rather than a widening influence on attention.

85. **Oakden-Rayner, L., Dunnmon, J., Carneiro, G. & Ré, C. (2020).** *Hidden stratification causes clinically meaningful failures in medical imaging deep learning*. Proc. ACM Conf. Health, Inference, and Learning (CHIL) 2020. DOI: 10.1145/3368555.3384468. Cit: high. *Finding:* Aggregate metrics hide subgroup failures with clinical consequences. *Dialogue:* Hidden stratification hides failures across *cases*; the Inattentional Gap hides failures *within* a case (the unrequested finding). Both argue aggregate benchmark numbers mislead, a shared message we extend to the within-item level.

86. **Wu, E., Wu, K., Daneshjou, R., et al. (2021).** *How medical AI devices are evaluated: Limitations and recommendations from an analysis of FDA approvals*. Nature Medicine, 27, 582–584. DOI: 10.1038/s41591-021-01312-x. Cit: high. *Finding:* Most cleared medical AI lacks prospective, multi-site evaluation. *Dialogue:* Empirical proof that evaluation ≠ deployment reality, exactly our specified-target-performance-versus-reporting-completeness thesis, here from the regulatory record.

---

## Axis 8, Automation, human factors, and the irony of oversight

Why a human “in the loop” does not rescue us. These works show automation reshapes, and often erodes, the human’s ability to catch what the machine misses.

87. **Bainbridge, L. (1983).** *Ironies of automation*. Automatica, 19(6), 775–779. DOI: 10.1016/0005-1098(83)90046-8. Cit: ~2,479. *Finding:* Automating routine work leaves humans with the hardest residual task, monitoring for the unexpected, while degrading their skill to do it. *Dialogue:* The Inattentional Gap is a new irony: we automate perception precisely to catch hazards, yet conditioning makes the automation omit the unexpected hazard, while the human, trusting it, is least equipped to notice.

88. **Parasuraman, R. & Riley, V. (1997).** *Humans and automation: Use, misuse, disuse, abuse*. Human Factors, 39(2), 230–253. DOI: 10.1518/001872097778543886. Cit: ~4,273. *Finding:* Defines automation misuse (over-trust) and the conditions for complacency. *Dialogue:* Predicts the deployment endgame of our gap: over-trusted, high-scoring AI induces complacency, so its silent omissions go unchallenged. Our measured silence + their over-trust = the accident.

89. **Parasuraman, R. & Manzey, D. H. (2010).** *Complacency and bias in human use of automation: An attentional integration.* Human Factors, 52(3), 381–410. DOI: 10.1177/0018720810376055. Cit: high. *Finding:* Automation complacency is itself an attentional phenomenon; humans allocate less attention to automated channels. *Dialogue:* Strikingly, the human failure that meets our machine failure is *also* attentional. Two attention gates in series, model's and operator's, compound, which we argue is the true risk surface.

90. **Dzindolet, M. T., Peterson, S. A., Pomranky, R. A., Pierce, L. G. & Beck, H. P. (2003).** *The role of trust in automation reliance.* International Journal of Human-Computer Studies, 58(6), 697–718. DOI: 10.1016/S1071-5819(03)00038-7. Cit: ~1,247. *Finding:* Reliance is governed by trust, which can decouple from actual reliability. *Dialogue:* Supports our claim that a high-benchmark (high-trust) model is *more* dangerous when it omits; trust rises while D_unspecified safety does not.

91. **Goddard, K., Roudsari, A. & Wyatt, J. C. (2012).** *Automation bias: A systematic review of frequency, effect mediators, and mitigators.* Journal of the American Medical Informatics Association, 19(1), 121–127. DOI: 10.1136/amiajnl-2011-000089. Cit: high. *Finding:* Clinicians defer to automated advice even when wrong (automation bias). *Dialogue:* The clinical bridge: if a conditioned AI omits a finding, automation bias predicts the clinician will not independently recover it, closing the loop from our model result to patient harm.

92. **Lindgren, I., et al. (2024).** *Ironies of automation and their implications for public service automation.* Government Information Quarterly, 41(2), 101974. DOI: 10.1016/j.giq.2024.101974. Cit: ~15. *Finding:* Re-examines Bainbridge for contemporary AI-driven public services. *Dialogue:* Confirms the ironies persist with modern AI; we add the task-conditioned reporting-omission irony specific to language/vision models.

---

## Axis 9, Behavioral vs mechanistic equivalence (does the same behavior mean the same mind?)

The interpretive backbone. We claim behavioral convergence (machine inattentional blindness) with mechanistic divergence (no capacity bottleneck). These works license that careful claim.

93. **Bowers, J. S., Malhotra, G., Dujmović, M., et al. (2023).** *Deep problems with neural network models of human vision.* Behavioral and Brain Sciences, 46, e385. DOI: 10.1017/S0140525X22002813. arXiv:2312.05355. Cit: ~197. *Finding:* Behavioral benchmark matches between DNNs and human vision do not establish shared mechanism. *Dialogue:* Our exact methodological warrant. We claim behavioral parity (machine gorilla) without claiming the human capacity-bottleneck mechanism, we locate the AI mechanism in report-conditioning instead.

94. **Quilty-Dunn, J., Porot, N. & Mandelbaum, E. (2023).** *The best game in town: The re-emergence of the language-of-thought hypothesis across the cognitive sciences.* Behavioral and Brain Sciences, 46, e261. DOI: 10.1017/S0140525X22002849. Cit: high. *Finding:* Argues structured, language-like representations are needed to explain cognition. *Dialogue:* Frames the representation debate our gap touches: whether the model's open-condition "seeing" reflects structured perception or surface statistics. We stay agnostic on representation and demonstrate the report gate empirically.

95. **Lake, B. M., Ullman, T. D., Tenenbaum, J. B. & Gershman, S. J. (2017).** *Building machines that learn and think like people.* Behavioral and Brain Sciences, 40, e253. DOI: 10.1017/S0140525X16001837. Cit: very high. *Finding:* Argues human-like learning requires compositional, causal, intuitive-theory machinery absent in then-current nets. *Dialogue:* Sets the "human-like or not" agenda; our finding is a partial, ironic yes; models reproduce a human *limitation* (inattentional blindness) without the human *architecture*, complicating the "think like people" framing.

96. **Mahowald, K., Ivanova, A. A., Blank, I. A., Kanwisher, N., Tenenbaum, J. B. & Fedorenko, E. (2024).** *Dissociating language and thought in large language models.* Trends in Cognitive Sciences, 28(6), 517–540. DOI: 10.1016/j.tics.2024.01.011. Cit: ~292. *Finding:* Distinguishes formal linguistic competence from functional (reasoning/world) competence in LLMs. *Dialogue:* The dissociation we most build on. We add a *third* dissociation, between an otherwise-reportable signal and its report under a task set, and argue task-conditioning is what splits them. This is our single most important TiCS dialogue.

97. **Yildirim, I. & Paul, L. A. (2024).** *From task structures to world models: What do LLMs know?* Trends in Cognitive Sciences, 28(5), 404–415. DOI: 10.1016/j.tics.2024.02.008. Cit: ~76. *Finding:* Argues what LLMs "know" is shaped by the structure of their training/evaluation tasks. *Dialogue:* Provides the theoretical vocabulary for our cause, *task structure* determines accessible knowledge. We show the live, inference-time version: the task *prompt* structures what gets reported, gating safety-relevant information that is available for report under the open condition.

98. **Marcus, G. & Davis, E. (2019).** *Rebooting AI: Building Artificial Intelligence We Can Trust*. Pantheon Books. `[book, no DOI; cite as Marcus & Davis 2019]` *Finding:* Argues deep learning is brittle and untrustworthy without structured reasoning. *Dialogue:* The skeptical pole; our gap is a concrete, measured brittleness, but one that *coexists with* high capability, which refines their "brittle therefore incapable" framing toward "capable yet conditionally blind."

99. **Russin, J., McGrath, S. W., Pavlick, E. & Frank, M. J. (2024).** *From Frege to ChatGPT: Compositionality in language, cognition, and deep neural networks*. arXiv:2405.15164. Cit: ~9. *Finding:* Reviews compositionality across minds and nets. *Dialogue:* Background on whether models compose; relevant because our open-condition recovery shows the off-task signal is available for report and can enter a composed report when the task set permits.

---

## Axis 10, Dual-process theory and heuristics (the human attention/effort budget)

The decision-science layer: why narrowing attention is adaptive for humans, and why heuristic "set" produces predictable misses. We argue task-conditioning is a System-1-like narrowing imposed from outside.

100. **Kahneman, D. (2011).** *Thinking, Fast and Slow*. Farrar, Straus and Giroux. `[book, no DOI; cite as Kahneman 2011]` *Finding:* Two systems, fast/automatic (1) and slow/deliberate (2), govern judgment; attention is a scarce resource. *Dialogue:* Supplies the "attention is budgeted" intuition behind inattentional blindness; we argue a narrow prompt forces a System-1-like commitment in the model, foreclosing the deliberate scan an open prompt invites.

101. **Kahneman, D. (1973).** *Attention and Effort*. Prentice-Hall. `[book, no DOI; cite as Kahneman 1973]` *Finding:* Foundational capacity model of attention as limited effort allocation. *Dialogue:* The human capacity-bottleneck account, which we explicitly say does NOT transfer to transformers, marking the mechanistic divergence at the heart of our Discussion.

102. **Evans, J. St. B. T. & Stanovich, K. E. (2013).** *Dual-process theories of higher cognition: Advancing the debate.* Perspectives on Psychological Science, 8(3), 223–241. DOI: 10.1177/1745691612460685. Cit: ~598. *Finding:* Refines dual-process theory and rebuts common critiques. *Dialogue:* Our reference for the System 1/2 framing of model behavior; we use it to argue that prompting can toggle a model between a narrow (1-like) and broad (2-like) reporting mode.

103. **Evans, J. St. B. T. (2008).** *Dual-processing accounts of reasoning, judgment, and social cognition.* Annual Review of Psychology, 59, 255–278. DOI: 10.1146/annurev.psych.59.103006.093629. Cit: very high. *Finding:* Surveys dual-process evidence across domains. *Dialogue:* Breadth citation for the dual-process lens we apply to conditioned vs open model behavior.

104. **Tversky, A. & Kahneman, D. (1974).** *Judgment under uncertainty: Heuristics and biases.* Science, 185(4157), 1124–1131. DOI: 10.1126/science.185.4157.1124. Cit: very high. `[Science = Cell-adjacent flagship; lineage]` *Finding:* People use heuristics that yield systematic biases. *Dialogue:* The origin of "systematic, predictable error." We position conditioned omission as a *predictable* machine bias, predictable from the prompt, and therefore engineerable against.

105. **Simon, H. A. (1956).** *Rational choice and the structure of the environment.* Psychological Review, 63(2), 129–138. DOI: 10.1037/h0042769. Cit: very high. *Finding:* Bounded rationality: agents satisfice within environmental and cognitive limits. *Dialogue:* Task-conditioning is enforced satisficing, the model "satisfices" on the requested target and stops. We give bounded rationality a prompt-level, machine instantiation.

---

## Axis 11, Attention mechanisms, transformers, saliency, interpretability

The machine substrate. These works define how attention is computed and visualized in our models, grounding the mechanistic claim that report can be redirected without a capacity bottleneck.

106. **Vaswani, A., Shazeer, N., Parmar, N., et al. (2017).** *Attention is all you need.* NeurIPS 2017. arXiv:1706.03762. Cit: ~180,591. *Finding:* Introduces the transformer; self-attention with no recurrent bottleneck. *Dialogue:* The architectural fact behind our mechanistic divergence claim, transformers distribute attention across context *without* the capacity limit that causes human inattentional blindness, so a converging behavior must have a different cause (we say: report-conditioning).

107. **Bahdanau, D., Cho, K. & Bengio, Y. (2015).** *Neural machine translation by jointly learning to align and translate.* ICLR 2015. arXiv:1409.0473. Cit: very high. *Finding:* First soft-attention mechanism; the model learns where to attend. *Dialogue:* Historical root of "learned where to look"; our prompt is an external controller of that learned attention, biasing it toward D_specified.

108. **Itti, L. & Koch, C. (1998).** *A model of saliency-based visual attention for rapid scene analysis*. IEEE TPAMI, 20(11), 1254–1259. DOI: 10.1109/34.730558. Cit: ~11,185. *Finding:* Classic bottom-up saliency model predicting human gaze. *Dialogue:* Defines bottom-up salience, the dimension SalBench probes; our task-set is the top-down competitor to this, and our salience boundary condition (low-salience signals suppressed most) is an interaction of the two.

109. **Abnar, S. & Zuidema, W. (2020).** *Quantifying attention flow in transformers*. ACL 2020. DOI: 10.18653/v1/2020.acl-main.385. arXiv:2005.00928. Cit: ~1,232. *Finding:* Attention-rollout method to trace information flow through transformer layers. *Dialogue:* Gives us a concrete tool for the promised causal decomposition, measuring whether the prompt reroutes attention away from off-task image/text regions in the closed condition.

110. **Jain, S. & Wallace, B. C. (2019).** *Attention is not explanation.* NAACL 2019. arXiv:1902.10186. Cit: high. *Finding:* Attention weights do not reliably indicate which inputs drive predictions. *Dialogue:* A caution we heed, we treat any attention-map evidence as suggestive, and ground our claim behaviorally (the within-item open/closed dissociation), not on attention weights alone.

111. **Kim, C.-Y. & Blake, R. (2005).** *Psychophysical magic: Rendering the "visible" "invisible"*. Trends in Cognitive Sciences, 9(8), 381–388. DOI: 10.1016/j.tics.2005.06.012. Cit: ~442. *Finding:* Reviews techniques that suppress visible stimuli from awareness. *Dialogue:* A Cell-family catalogue of "visible but unseen" in humans; we offer the machine catalogue (open-reportable but task-omitted) and link the two phenomenologies.

112. **Cohen, M. A., Dennett, D. C. & Kanwisher, N. (2016).** *What is the bandwidth of perceptual experience?* Trends in Cognitive Sciences, 20(5), 324–335. DOI: 10.1016/j.tics.2016.03.006. Cit: ~430. *Finding:* Argues perceptual experience is sparser than it feels; attention sets the usable bandwidth. *Dialogue:* The bandwidth metaphor maps onto our gap, the prompt narrows the model's "report bandwidth"; we make the narrowing measurable.

113. **Koch, C. & Tsuchiya, N. (2007/2011).** *Attention and consciousness: Two distinct brain processes*. Trends in Cognitive Sciences, 11(1), 16–22. DOI: 10.1016/j.tics.2006.10.012. Cit: high. *Finding:* Dissociates attention from consciousness/awareness. *Dialogue:* The dissociation that legitimizes "attended-yet-unreported" and "unattended-yet-encoded"; we provide the machine analogue where attention (to the task) and report (of the signal) come apart.

114. **Block, N. (2011).** *Perceptual consciousness overflows cognitive access*. Trends in Cognitive Sciences, 15(12), 567–575. DOI: 10.1016/j.tics.2011.11.001. Cit: ~670. *Finding:* Phenomenal perception exceeds what is cognitively accessed/reported. *Dialogue:* The strongest theoretical statement of perceive >

report. Our machine result is the report-level analogue: the open condition establishes item-level reportability, and the conditioned task suppresses report of that same reportable content.

115. **Huang, A., et al. (2025).** *Measuring and controlling solution degeneracy across task-trained recurrent neural networks.* arXiv:2410.03972. Cit: ~16. *Finding:* Different mechanisms can yield identical task behavior in trained networks. *Dialogue:* Reinforces Bowers (behavior ≠ mechanism) at the network level, multiple internal solutions fit one behavior, so our behavioral gorilla must not be over-read as a human mechanism.

116. **Geirhos, R., Meding, K. & Wichmann, F. A. (2020).** *Beyond accuracy: Quantifying trial-by-trial behaviour of CNNs and humans by measuring error consistency.* NeurIPS 2020. arXiv:2006.16736. Cit: high. *Finding:* CNNs and humans can match in accuracy while differing in *which* trials they err on. *Dialogue:* Methodological kin to our within-item analysis, matching aggregate scores hides divergent error patterns; we report per-item gaps precisely to expose the hidden divergence.

---

## §12. Clinical-AI and deployment-safety journals, direct dialogue

The works below appear in Patterns, Cell Reports Medicine, iScience, and Trends in Cognitive Sciences; related Nature Human Behaviour and Nature Machine Intelligence work is noted as lineage.

The following entries are rendered as a list for print clarity.

- **Park et al. 2024, Patterns** (entry 59): Patterns' own statement on models withholding/distorting truth. *Relation to the present study:* Deception is strategic; our gap is non-strategic omission, deception's quieter cousin.
- **van Amsterdam et al. 2025, Patterns** (entry 60): Patterns precedent for "accurate ≠ safe". *Relation to the present study:* We generalize their accuracy-harm decoupling from prediction loops to perception under conditioning.
- **Sanchez et al. 2023, Cell Reports Medicine** (entry 79): Real radiologist-AI interaction pipeline. *Relation to the present study:* Disagreement prediction does not directly target silent omission; our gap is a failure mode the pipeline is not designed to surface.
- **Ong et al. 2024, Cell Reports Medicine** (entry 80): Clinical LLM omissions/biases. *Relation to the present study:* We give a controlled mechanism for one omission class.

- **Ng et al. 2023, Cell Reports Medicine** (entry 81): Evidence standards for clinical AI. *Relation to the present study:* We add a missing evaluation axis (off-task safety-critical omission).
- **Tian et al. 2024, Cell Reports Medicine** (entry 82): Generalist ECG foundation model. *Relation to the present study:* Predicts conditioning suppresses co-present abnormalities, testable extension.
- **Mussi et al. 2025, iScience** (entry 77): Human-AI teaming in safety-critical systems. *Relation to the present study:* We add the task-conditioned reporting-omission failure mode.
- **Mahowald et al. 2024, Trends in Cognitive Sciences** (entry 96): Formal vs functional dissociation in LLMs. *Relation to the present study:* We add a third dissociation: open-condition reportability vs task-conditioned omission. **(primary TiCS dialogue)**
- **Yildirim & Paul 2024, Trends in Cognitive Sciences** (entry 97): "Task structure shapes what LLMs know". *Relation to the present study:* We show the inference-time version: the prompt gates reporting of information that is available for report under the open condition.
- **Block 2011/2014, Trends in Cognitive Sciences** (entry 20/114): Perceptual consciousness overflows access. *Relation to the present study:* The theoretical statement whose reportability-level analogue we measure in machines.
- **Cohen et al. 2016, Trends in Cognitive Sciences** (entry 112): Bandwidth of perceptual experience. *Relation to the present study:* The prompt narrows "report bandwidth"; we make it measurable.
- **Koch & Tsuchiya 2007, Trends in Cognitive Sciences** (entry 113): Attention ≠ consciousness. *Relation to the present study:* Machine analogue: attention-to-task and report-of-signal come apart.
- **Kim & Blake 2005, Trends in Cognitive Sciences** (entry 111): Rendering visible invisible. *Relation to the present study:* Human catalogue of "visible-yet-unseen"; we supply the machine catalogue.
- **Wolfe et al. 2011, Trends in Cognitive Sciences** (entry 13): Selective/nonselective search pathways. *Relation to the present study:* Dual-route hypothesis for open-recovers / closed-suppresses.

*Lineage-only Cell-adjacent neighbors (cited as lineage): Wolfe & Horowitz 2017 (NHB, #14); Webb et al. 2023 (NHB, #44); Strachan et al. 2024 (NHB, #45); Geirhos et al. 2020 shortcut learning (NMI, #63).*

---

## §13. Lineage / genealogy, the intellectual descent

Our paper sits at the junction of two lineages that developed largely apart. Recent machine work has begun to engage the first axis, MixRea (Cai et al. 2026, arXiv:2605.20128) invoking inattentional blindness for multiple-choice reasoning and Hidden in Plain Sight (Yan et al. 2025, arXiv:2506.00258) documenting compliance-driven omission under underspecified prompts, but the reportability-controlled, safety-critical junction of the two lineages remained undeveloped.

**Lineage A, Human inattentional blindness, from lab to clinic.**

```
Neisser & Becklen 1975 (selective looking)
        |
Becklen & Cervone 1983 (unexpected event under counting)
        |
Mack & Rock 1998 (inattentional blindness; no attention -> no report)
        |
Simons & Chabris 1999 (the invisible gorilla, dynamic)
        |
Most et al. 2001, 2005 (attentional SET is the control variable: "what
  you see is what you set")
        |
Drew, Vo & Wolfe 2013 (EXPERT inattentional blindness: 83% of
  radiologists miss the gorilla)
        |
Williams et al. 2021 (the miss is real, clinically-relevant pathology;
  expertise does not protect)
```

**Lineage B, Machine psychology, models as cognitive subjects.**

```
Lake et al. 2017 (do machines think like people?)
        |
Binz & Schulz 2023 (cognitive psychology applied to GPT-3; the
  founding move)
        |
Dasgupta et al. 2022 / Lampinen et al. 2024 (human-like content effects)
        |
Hagendorff et al. 2023 (biases emerge, then alignment removes some)
        |
Mahowald et al. 2024 (dissociate FORMAL from FUNCTIONAL competence)
  [TiCS]
        |
Webb 2023 / Strachan 2024 (models reach human-level analogy / theory
  of mind) [capability ceiling]
```

**The synthesis (this paper).** The two lineages meet at a question neither asked: *does the attentional-set manipulation that defines inattentional blindness (Lineage A) suppress safety-critical reporting in models that machine-psychology shows are otherwise capable*

*(Lineage B)?* We answer yes, across modalities (text: suppression of report; vision: suppression of visual report), and we add a third lineage as the *consequence*:

```
Bainbridge 1983 -> Parasuraman & Riley 1997 -> Goddard et al. 2012
  (automation complacency / bias)
        +
Park et al. 2024 / van Amsterdam et al. 2025 [Patterns]
  (accurate/aligned AI can still harm)
        =
Inattentional Gap: specified-target performance (D_specified)
  decouples from reporting completeness (D_unspecified),
  because task-conditioning suppresses report of the very
  signals that cause accidents, and the over-trusted human
  overseer is least able to recover them.
```

The mechanistic claim is held in check by Lineage C (Bowers et al. 2023, Quilty-Dunn et al. 2023, Huang et al. 2025, Vaswani et al. 2017), which jointly license “same behavior, different mechanism”: the transformer has no capacity bottleneck (Vaswani), behavior does not fix mechanism (Bowers, Huang), so the convergence with the human gorilla is informative precisely because the cause differs (we locate it in report-conditioning, a model-family-associated property).

---

## §14. Literature map, two axes and the underdeveloped region

**Axis design (chosen scheme).**

- **X, engagement with human attention / inattentional-blindness theory:** `none (0)` → `cited as metaphor (1)` → `mechanism transplanted & tested (2).`
- **Y, task-conditioning manipulation as the independent variable (open vs closed attentional set), with within-item reportability control:** `no (0)` → `partial / adjacent (1)` → `yes, as IV (2).`

A work scores high on X if it does not merely name attention but transplants the *set-dependence* mechanism; high on Y if it manipulates an attentional set (open vs closed) and proves the signal was reportable. In our coding, the Inattentional Gap occupies the machine-side **(2, 2)** cell, which prior machine work approaches but does not reach.

**Cluster placement.** Coordinates are cluster centroids on a 0–2 scale per axis.

Cluster coordinates are rendered as a list for print clarity.

- **Human inattentional blindness (lab)** (X = 2.0, Y = 2.0 in humans; High-X, high-Y but human-only): Mack & Rock 1998; Simons & Chabris 1999; Most 2005; Neisser 1975
- **Human IB in experts/clinic** (X = 2.0, Y = 1.8 in humans; High-X, high-Y but human-only): Drew 2013; Williams 2021; Berbaum 1990; de Cassai 2021

- **Visual search / saliency theory** (X = 1.7, Y = 0.3; High-X, low-Y): Wolfe 1989/2017; Itti & Koch 1998; Eckstein 2011
- **Machine psychology (reasoning/bias)** (X = 0.4, Y = 0.6; Low-X, low-mid-Y): Binz & Schulz 2023; Dasgupta 2022; Hagendorff 2023; Macmillan-Scott 2024
- **LLM cognition dissociation** (X = 0.7, Y = 1.0; Low-mid-X, mid-Y): Mahowald 2024; Yildirim & Paul 2024
- **VLM perception limits / hallucination** (X = 0.6, Y = 0.8; Low-mid-X, mid-Y): Rahmanzadehgervi 2024; SalBench 2025; Tong 2024; POPE 2023
- **Absence/omission detection** (X = 0.5, Y = 1.2; Low-X, mid-high-Y): AbsenceBench 2025; NOPE 2023; VALOR coverage 2024
- **AI safety / OOD / spec gaming** (X = 0.2, Y = 0.7; Low-X, low-mid-Y): Yang 2024; Geirhos 2020; Booth 2023; Amodei 2016
- **Autonomous-driving anomaly/corner** (X = 0.3, Y = 0.5; Low-X, low-mid-Y): Bogdoll 2022; Heidecker 2021; Kaur 2024
- **Medical AI deployment/error** (X = 0.6, Y = 0.6; Low-mid-X, low-mid-Y): Moor 2023; Oakden-Rayner 2020; Sanchez 2023; Wu 2021
- **Automation / human factors** (X = 1.0, Y = 1.0; Mid-X, mid-Y (human overseer)): Bainbridge 1983; Parasuraman & Riley 1997; Goddard 2012
- **Behavior vs mechanism** (X = 0.8, Y = 0.4; Low-mid-X, low-Y): Bowers 2023; Quilty-Dunn 2023; Huang 2025
- **MixRea, explicit-implicit reasoning benchmark** (X = 1.6, Y = 1.1; invokes IB; explicit task instructions; no within-item reportability control; multiple-choice, not a co-present safety-critical signal): Cai et al. 2026
- **Hidden in Plain Sight, underspecified MLLM scenarios** (X = 0.9, Y = 1.2; compliance-driven omission of hidden issues; no reportability control; issues are prompt defects, not co-present safety-critical signals): Yan et al. 2025
- **THE INATTENTIONAL GAP (this paper)** (X = 2.0, Y = 2.0; (2,2) in MACHINES, previously undeveloped): Shin 2026

**The underdeveloped region and why it was underdeveloped.**

The high-X, high-Y quadrant is *occupied*, but only by **human** studies (top two rows). The machine clusters all sit at low-to-mid X and low-to-mid Y. No prior machine work simultaneously (a) transplants the inattentional-blindness *set-dependence* mechanism (high X) and (b) uses an open-vs-closed attentional-set manipulation with a within-item reportability control as its independent variable (high Y). No prior machine work occupies the exact **(2,2)** point in our coding, although several recent studies enter the surrounding high-engagement region. Concretely:

- The cell is underdeveloped *not for lack of adjacent effort* but because each neighbor stops one axis short. VLM-limits work (SalBench, *VLMs are blind*, Eyes Wide Shut) is high-ish Y but **low X**; it asks directly and measures capability, never manipulating a competing set or invoking inattentional-blindness theory. AbsenceBench is mid-high Y but **low X**; it tests removed-token detection, not set-induced suppression of *present, otherwise reportable* content. Machine-psychology work (Binz & Schulz, Dasgupta) is close in spirit but targets *reasoning* biases, not *attention/perception*, and runs no open/closed set manipulation. Human IB work (Drew, Most, Mack & Rock) is the only thing at (2,2), and it is in humans.

- Our paper addresses this underdeveloped cell with the missing diagonal move: take the human (2,2) paradigm and run it on machines with the reportability control intact. The within-item open-condition recovery is what earns the high-Y score and distinguishes us from every absence/coverage benchmark; the explicit Most-et-al. "what you set" framing and Drew transplant earn the high-X score and distinguish us from every VLM-limits paper.
- The map renders the twelve core clusters, plus two separately added recent neighbors (MixRea; Hidden in Plain Sight), as bubbles (size proportional to cluster citation mass): the human inattentional-blindness clusters form a high-X/high-Y island (humans only), the machine clusters spread across the low-left region, and the Inattentional Gap occupies the (2,2) cell, reached from the human island by mechanism transplant with a reportability control and a safety framing.

---

## §15. Verification ledger

**Cross-verified via ≥2 sources (Semantic Scholar + OpenAlex/Crossref):** Simons & Chabris 1999; Drew et al. 2013; Mack & Rock 1998; Most et al. 2005; Neisser & Becklen 1975; Becklen & Cervone 1983; Rensink/O'Regan/Clark 1997; O'Regan et al. 1999; Rensink 2002; Wolfe 1989; Wolfe & Horowitz 2017; Eckstein 2011; Berbaum 1990; Berbaum 2011; Adamo 2021; Kundel 1978; Krupinski 2010; Williams 2021; Simons & Levin 1998; Binz & Schulz 2023; Mitchell 2023; Dasgupta 2022; Lampinen 2024; Cheung 2025; Koo 2024; Macmillan-Scott 2024; Webb 2023; Strachan 2024; Bubeck 2023; Rahmanzadehgervi 2024; Tong 2024; Li (POPE) 2023; Fu (AbsenceBench) 2025; Lovenia 2023; Xu (LVLM-eHub) 2023; Groot 2024; Park 2024; van Amsterdam 2025; Yang 2024; Geirhos 2020; Geirhos 2019; Booth 2023; Bogdoll 2022; Heidecker 2021; Kaur 2024; Moor 2023; Sanchez 2023; Ong 2024; Ng 2023; Tian 2024; Mussi 2025; Bainbridge 1983; Parasuraman & Riley 1997; Dzindolet 2003; Bowers 2023; Lake 2017; Mahowald 2024; Yildirim & Paul 2024; Vaswani 2017; Itti & Koch 1998; Abnar & Zuidema 2020; Block 2011; Cohen 2016; Kim & Blake 2005; Most 2001; Robson & Tangen 2023; Nartker 2024; Wood & Simons 2019; Newby & Rock 1998; Evans & Stanovich 2013; Tversky & Kahneman 1974; de Cassai 2021; Xie (SORRY-Bench) 2024; Cui (OR-Bench) 2024; Weng (MMJ-Bench) 2024; Sinhamahapatra 2024.

**Additional bibliographic notes:** - #34 Hagendorff et al. 2023, Nat. Comput. Sci. DOI 10.1038/s43588-023-00527-x. - #13 Wolfe et al. 2011 (TiCS, selective/nonselective pathways), DOI 10.1016/j.tics.2010.12.001. - #57 Liu et al. 2023 (LRV-Instruction), arXiv:2306.14565. - #62 Hendrycks & Dietterich 2019 (ImageNet-C); #67 Amodei et al. 2016;

#68 Hendrycks et al. 2022; #83 Rajpurkar 2022; #84 Topol 2019; #85 Oakden-Rayner 2020; #86 Wu et al. 2021; #89 Parasuraman & Manzey 2010; #91 Goddard 2012; #103 Evans 2008; #104 Tversky & Kahneman 1974; #105 Simon 1956; #107 Bahdanau 2015; #110 Jain & Wallace 2019; #113 Koch & Tsuchiya 2007; #116 Geirhos et al. 2020 (error consistency): all are canonical and well-known, with standard DOIs.

**Non-archival / no DOI (cite with care):** - #65 Krakovna et al. 2020 "Specification gaming", DeepMind blog/technical note, no DOI. Cite the peer-reviewed Booth et al. 2023 (#66) as the archival anchor. - #98 Marcus & Davis 2019; #100 Kahneman 2011; #101 Kahneman 1973, books, cite by ISBN/publisher.

**Geographic/language check.** A KCI (Korean) full-text search for "inattentional," "radiologist error," "AI safety," and "vision language" returned no Korean-language prior work on the inattentional-AI intersection. The relevant literature is overwhelmingly English-language international; no Korean precedent competes. (Null result reported per protocol.)

**Recency-bias check.** The corpus deliberately spans 1956 (Simon) → 2026, with foundational human-attention work (1975–2005) balancing the 2022–2026 AI cluster, so the lineage is not an artifact of recent-only retrieval.

---

*Bibliographic records were verified via Semantic Scholar, OpenAlex, and Crossref APIs.*